\documentclass[lettersize,journal]{IEEEtran}

\ifCLASSINFOpdf
  \usepackage[pdftex]{graphicx}
  \graphicspath{{../pdf/}{../jpeg/}}
  \DeclareGraphicsExtensions{.pdf,.jpeg,.png}
\else
  \usepackage[dvips]{graphicx}
  \graphicspath{{../eps/}}
  \DeclareGraphicsExtensions{.eps}
\fi

\usepackage{amsmath,amsfonts}
\usepackage{algorithmic}
\usepackage{algorithm}
\usepackage{array}
\usepackage[caption=false,font=normalsize,labelfont=sf,textfont=sf]{subfig}
\usepackage{textcomp}
\usepackage{stfloats}
\usepackage{url}
\usepackage{tablefootnote}
\usepackage{verbatim}
\usepackage{graphicx}
\usepackage{xcolor}
\usepackage{multirow}
\usepackage{cite}
\begin{document}

\title{Weakly Supervised Learning for Facial \\Affective Behavior Analysis: A Review}

\author{R. Gnana Praveen,~
Patrick~Cardinal,~\IEEEmembership{Member,~IEEE}
        and~Eric Granger,~\IEEEmembership{Member,~IEEE}
\thanks{R. Gnana Praveen is with Huawei Noah's Ark Lab, Montreal, Canada. Patrick Cardinal and Eric Granger are with Laboratoire d'imagerie, de vision et d'intelligence artificielle (LIVIA), 
ETS Montreal, Canada.  
E-mail: praveenrgp1988@gmail.com, \{patrick.cardinal, eric.granger\}@etsmtl.ca.}
}

\markboth{Journal of \LaTeX\ Class Files,~Vol.~14, No.~8, August~2021}%
{Shell \MakeLowercase{\textit{et al.}}: A Sample Article Using IEEEtran.cls for IEEE Journals}

\maketitle
\begin{abstract}
Recent advances in deep learning (DL) and computational capacity have enabled facial affective behavior analysis (FABA) to progress from static images captured in controlled settings to fine-grained analysis of facial expressions in real-world video data. However, training accurate DL models for FABA typically requires large-scale, expert-annotated datasets, which are costly to obtain and inherently noisy due to the ambiguity of labeling subtle facial expressions and action units (AUs). To mitigate these challenges, weakly supervised learning (WSL) has emerged as a promising paradigm for training models with weak annotations. In this paper, we present a structured taxonomy of WSL scenarios for FABA, organized according to the type of weak annotation and the specific affective task. Building on this taxonomy, we provide a critical synthesis of representative WSL methods for both classification (expression and AU recognition) and regression (expression and AU intensity estimation) tasks, focusing on their core methodological ideas, strengths, and limitations. Furthermore, we systematically summarize the comparative performance of WSL approaches along with widely adopted experimental setups and evaluation protocols. Our critical assessment identifies key challenges and future research directions, including the need for efficient adaptation of foundation models and for the development of robust, scalable FABA systems suitable for real-world applications. 
\end{abstract}


\section{Introduction}

\begin{figure}[!t]
\centering
\includegraphics[width=0.45\textwidth]{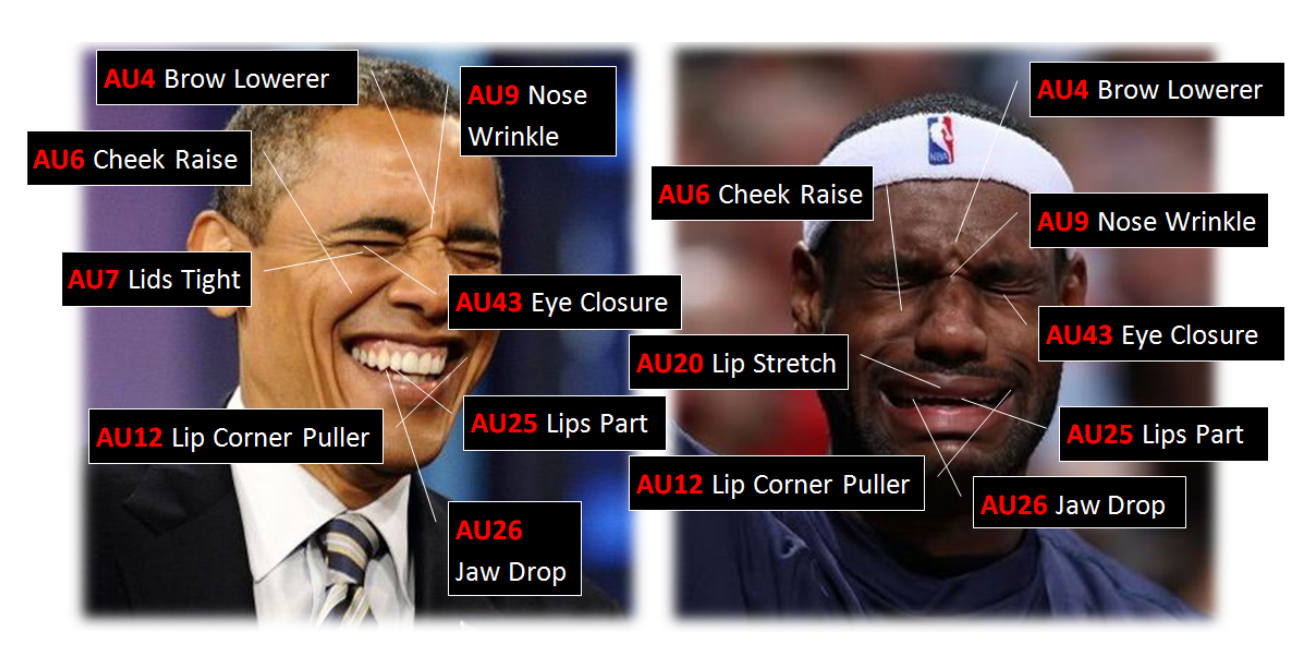}
\caption{Examples of facial images with action units (AUs) \cite{Martinez2016}.}
\label{fig:AU}
\vspace{-4mm}
\end{figure}

\begin{figure*}[!t]
\centering
\includegraphics[width=0.9\textwidth]{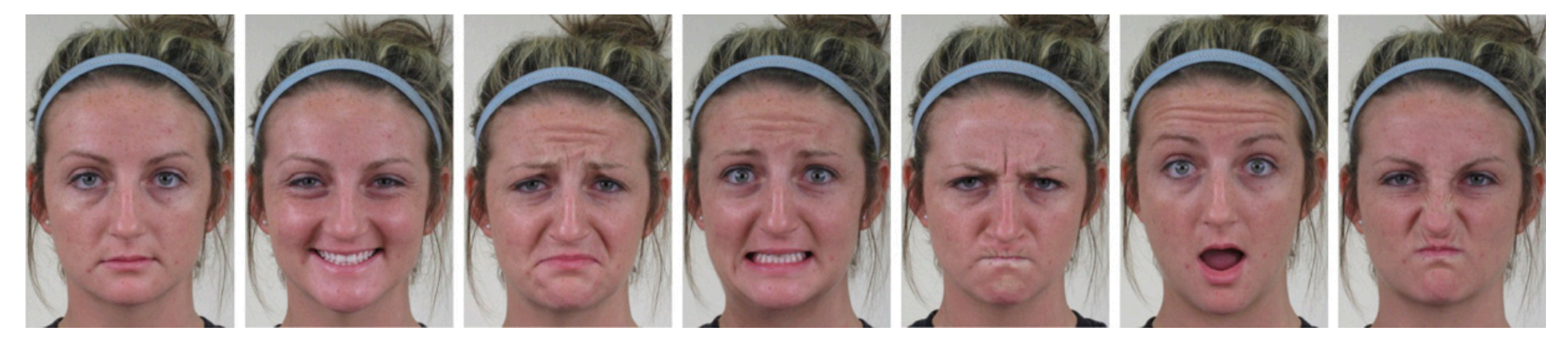}
\caption{Examples of primary universal emotions. From left to right: neutral, happy, sad, fear, angry, surprise, and disgust. Images are taken from the Compound Facial Expressions of Emotions dataset \cite{FE}.}
\label{fig:FE}
\end{figure*}

\IEEEPARstart{F}{acial} affective behavior analysis (FABA) plays a vital role in computer vision and affective computing, enabling a wide range of applications such as human-computer interaction \cite{10.1145/3549865.3549908}, medical diagnosis \cite{de2020deep}, e-education \cite{Thomas:2018:PEI:3242969.3264984}, and autonomous driving \cite{SAADI2024122784}. 
Although affective cues can be conveyed through multiple modalities, facial behavior manifested in the intricate movements of facial muscles provides an exceptionally rich wealth of information, making it invaluable for accurate analysis of affective states \cite{Mehrabian}. To leverage this information, three complementary annotation paradigms have emerged, ranging from coarse categorical labels to fine-grained intensity and dimensional scores. First, the categorical model defines six basic universal emotions—anger, disgust, fear, happiness, sadness, and surprise \cite{ekman76} (see Fig. \ref{fig:FE}). Subsequently, contempt was added as a seventh category \cite{Matsumoto1992}. This framework offers high-level interpretability of affective states and is widely used for classification tasks. In addition to that, micro-expressions and compound expressions are also introduced to capture the mixed emotional states \cite{FE,9915437}. 
Second, the facial action coding system (FACS) \cite{Ekman1978} provides a more granular low-level muscle-based representation by defining action units (AU), each corresponding to the movement of specific facial muscles as shown in Fig. \ref{fig:AU}. 
Finally, beyond these classification schemes, regression-based annotations are also explored to capture the intensity of affective states based on ordinal intensity levels or dimensional modeling. Ordinal regression represents expressions or AUs in discrete intensity levels, whereas dimensional modeling maps affect to continuous scales, often using valence (unpleasant to pleasant) and arousal (calm to excited) to capture the nuanced and fine-grained emotional states \cite{POSNER_RUSSELL_PETERSON_2005}.  

Despite their widespread use, obtaining high-quality annotations remains a significant challenge, as the ambiguity increases when transitioning from coarse categorical labels to more nuanced, intensity-based, or continuous representations of emotions\footnote{Though the terms ”expression” and ”emotion” are used interchangeably in the literature, the primary difference is that facial emotion conveys the mental state of a person, whereas facial expression is the indicator of the emotions being felt, i.e., facial expressions display a wide range of facial modulations, but facial emotions are limited.}. Categorical annotations, such as discrete emotion labels are relatively straightforward to obtain and interpret but fail to capture the fine-grained emotional states.  
Although FACS-based AU annotations are more granular and informative, they require substantial expertise and manual effort. Achieving certification as a FACS coder requires over 100 hours of training, and annotating even a single minute of video can take close to an hour \cite{Ekman1978}. Assessing AU intensities introduces further ambiguity, as low-intensity or subtle facial muscle activations are difficult to quantify consistently. Dimensional annotations, such as continuous valence and arousal ratings, are even more challenging, requiring sustained affective judgments over time that are cognitively taxing, prone to annotator fatigue, and often yield low inter-rater reliability.

Given these challenges in obtaining high-quality annotations, training deep learning (DL) models for FABA remains difficult, as these models are both data-intensive and highly sensitive to label noise. In practice, collecting large-scale, fully annotated datasets is often infeasible, particularly for application-specific or fine-grained affective tasks. To address these limitations, two broad research paradigms have been explored: (1) unsupervised or self-supervised learning \cite{ZHANG2025131409} and (2) weakly supervised learning \cite{WSL}. While self-supervised learning offers a promising alternative, they require massive pretraining datasets and may not directly capture task-specific affective cues without additional supervision. On the other hand, WSL offers a middle ground instead of discarding supervision altogether; it leverages weak labels such as imprecise or incomplete labels (e.g., video-level emotion labels) to guide the learning process. This paradigm aligns closely with the practical realities of affective computing, where obtaining perfect labels is unrealistic, but some form of weak supervision is often available.
In this paper, we examine how WSL approaches can mitigate the challenges of annotation bottlenecks to build robust FABA systems by leveraging weak annotations of expressions and AUs. Note that this survey excludes papers that rely only on self-supervised learning models without any external supervision.  

Existing surveys on FABA systems focus primarily on fully supervised methods \cite{Martinez2016, 6940284, FASEL2003259,9915437, CANAL2022593}, including deep facial expression recognition (FER) \cite{li2020deep}, or specific tasks such as pain estimation \cite{8928510}, often assuming the availability of densely annotated datasets. Motivated by the growing importance of addressing annotation bottlenecks in real-world scenarios, this survey presents a comprehensive and critical review of state-of-the-art WSL methods developed for FABA systems. We systematically identify and organize the core sub-problems in WSL for FABA, including classification with weak categorical or discrete labels, and regression with weak ordinal or dimensional annotations. For each scenario, we analyze representative methods by summarizing their core ideas and insights, as well as their advantages and limitations. Additionally, we examine the datasets and evaluation protocols commonly used in WSL scenarios of FABA systems, including training strategies and performance trade-offs. Based on our critical analysis of existing methods, we discuss open challenges and future research directions aimed at building robust, data-efficient FABA systems, including adapting foundation models and few-shot learning.

The main contributions of this paper can be summarized as: (1) A structured taxonomy of WSL scenarios for FABA, highlighting their relevance across different annotation paradigms. This taxonomy formulates problem settings based on the type of weak annotation and the specific affective task, and identifies the key challenges associated with each scenario 
(Section \ref{WSL}). (2) A comprehensive and critical review of state-of-the-art WSL methods proposed under different WSL scenarios for classification (i.e., expression and AU recognition), and for regression (i.e., expression and AU intensity estimation), along with the critical analysis of their strengths and limitations (Sections \ref{WSL for classification} and \ref{WSL for Regression}, respectively). (3) An overview of experimental setups and benchmarking protocols, and comparative performance analysis of these methods under different WSL scenarios (Sections \ref{exp for classification} and \ref{exp for ordinal regression}). \noindent (4) A discussion on open challenges and future research directions, with an emphasis on scalable and reliable FABA systems under real-world weakly annotated data (Section \ref{Challenges and Opportunities}).

\section{Datasets} 
\label{Datsets}


In this section, we review widely used datasets for evaluating WSL methods under different forms of weak annotation in FABA. These datasets vary in annotation granularity, supervision level, and affective task, and are often adapted to specific WSL scenarios using standardized evaluation protocols.

\noindent \textbf{UNBC-McMaster \cite{5771462}: }
This database contains 200 video sequences of 48398 frames captured from 25 participants experiencing shoulder pain. Each frame of the video sequence is labeled with discrete intensity levels (A–E) of pain-related AUs (AU4, AU6, AU7, AU9, AU10, AU12, AU20, AU25, AU26, AU27 and AU43) by certified FACS coders. 
In addition, the dataset provides discrete pain intensity labels at both the frame and sequence levels, using the Prkachin and Solomon Pain Intensity Scale (PSPI; 0–15) and the Observer Pain Intensity (OPI; 0–5).



\noindent \textbf{DISFA \cite{6475933}: }
The dataset is created using 9 short video clips from YouTube, where 27 adults are allowed to watch the short video clips about various emotions. The facial expressions of each of the participants are captured with a high-resolution video of 1024x768 pixels with a frame rate of 20fps resulting in 1,30,754 frames. Each of these frames is annotated with AUs along with the discrete intensity levels by FACS expert raters. The AUs related to the expressions in the database are AU1, AU2, AU4, AU5, AU6, AU9, AU12, AU15, AU17, AU20, AU25, and AU26, whose intensities are provided on a six-point ordinal scale (neutral $<$ A $<$ B $<$ C $<$ D $<$ E). 

\noindent \textbf{BP4D \cite{ZHANG2014692}: }
The dataset was collected from 41 participants, each one asked to exhibit eight spontaneous facial expressions, resulting in the acquisition of both 2D and 3D video recordings for each task. In total, 328 video sequences were obtained. Due to the high cost and complexity of FACS-based annotation, only the most expressive temporal segments—specifically, 20-second segments with high facial activity—were annotated with facial action units (AUs). Across these sequences, 27 AUs were coded for expression analysis, with ordinal intensity annotations (0–5) provided for AU12 and AU14. 


\noindent \textbf{FER+ \cite{Barsoum:2016:TDN:2993148.2993165}:}
The dataset is an extension of the FER2013 dataset \cite{FER2013}, which is labeled automatically by the Google image search engine. All the images in FER2013 have been registered and resized to 48x48 pixels. The dataset is partitioned into 28,709 training images, 3,589 validation images, and 3,589 test images with seven basic expression labels. The images are further relabeled by 10 individuals, thereby obtaining more reliable annotations.     

\noindent \textbf{CK+ \cite{5543262} :}
This dataset comprises 593 video sequences captured under controlled laboratory conditions, in which 123 participants spontaneously performed facial expressions. Each sequence depicts a progression from a neutral facial state to the peak expression, with sequence lengths ranging from 10 to 60 frames. Because the original emotion labels were found to be unreliable, the annotations were subsequently refined using the FACS to assign seven basic expression categories, including contempt. Each expression category is defined by a prototypical combination of specific action units (AUs). Following this refinement process, 327 sequences were retained, with each sequence labeled at the video level with a single expression category. For static image-based approaches, the video-level label is commonly associated with the last one to three frames, corresponding to the peak expression, while treating the initial frame as neutral.

\noindent \textbf{MMI \cite{1521424}:}
The database contains 326 video sequences spontaneously captured in laboratory-controlled conditions from 32 subjects, and it includes challenging variations such as interpersonal variations, pose, etc., compared to the CK+ database. The sequences are captured as onset-apex-offset, i.e., the sequence starts with neutral expression (onset), reaches the peak (apex), and returns to neutral expression (offset). As per the standard of FACS, 213 sequences are labeled with six basic facial expressions (excluding contempt) at the video level, of which 205 sequences are captured in frontal view. They have also provided frame-level annotations for some of the sequences. For approaches based on static images, only the first frame (neutral expression) and peak frames (apex of expressions) are considered.

\noindent \textbf{RAF-DB \cite{8099760}: }
This large-scale in-the-wild dataset consists of extremely diverse facial images downloaded from the Internet. The obtained images were labeled by 315 annotators, and each image was ensured to be labeled by 40 independent annotators. The final annotations were obtained using crowd-sourcing techniques for seven basic emotions (including neutral expression). The dataset is partitioned into 12,271 training samples and 3,068 test samples. Some of the images are also labeled with compound expressions of 11 classes.    

\noindent \textbf{AffectNet \cite{8013713}: }
This is the largest dataset captured in the wild, consisting of 0.4 million images, downloaded from the Internet using three search engines based on expression-related keywords. The images are labeled with seven basic expressions (excluding neutral), and partitioned to 2,80,000 samples for training and 3,500 images for validation.

\noindent \textbf{FERA 2015 Challenge \cite{7284874}: } 
The dataset was constructed from the BP4D \cite{ZHANG2014692} and SEMAINE \cite{5959155} datasets for AU occurrence detection and AU intensity estimation. For intensity estimation, only five AUs from BP4D are considered—AU6, AU10, AU12, AU14, and AU17—while AU occurrence detection is performed on a larger set of 14 AUs shared across BP4D and SEMAINE (AU1, AU2, AU4, AU6, AU7, AU10, AU12, AU14, AU15, AU17, AU23, AU25, AU28, and AU45). For BP4D, the original dataset is partitioned into subject-independent splits, with 21 subjects used for training, 20 for development, and an additional 20 subjects forming an extended test set. This results in 75,586 images for training, 71,261 for development, and 75,726 for testing. Similarly, the SEMAINE dataset comprises 48,000 training images, 45,000 development images, and 37,695 testing images. Both datasets provide frame-level annotations for AU occurrence and, where applicable, AU intensity. In the BP4D-extended set, onset and offset frames are assigned a B-level (level 2) intensity to maintain consistency with the ordinal labeling scheme. Across both datasets, most facially expressive segments are annotated, and AU intensities are encoded on an ordinal scale from 0 to 5, enabling fine-grained analysis of expressions.


\noindent \textbf{MAFW \cite{10.1145/3503161.3548190}: } This is the only data set consisting of short textual descriptions of facial expressions in two languages, English and Mandarin Chinese, along with categorical annotations of 11 categories of anger, disgust, fear, happiness, neutral, sadness, surprise, contempt, anxiety, helplessness, and disappointment in addition to 32 compound expression categories. The dataset provides a large-scale multi-modal compound affective database with 10,045 video-audio clips in the wild. Each clip is annotated with an emotional category and a couple of sentences describing the subject's affective behaviors in the clip. 


\begin{figure*}[!t]
\centering
\includegraphics[width=1.0\textwidth]
{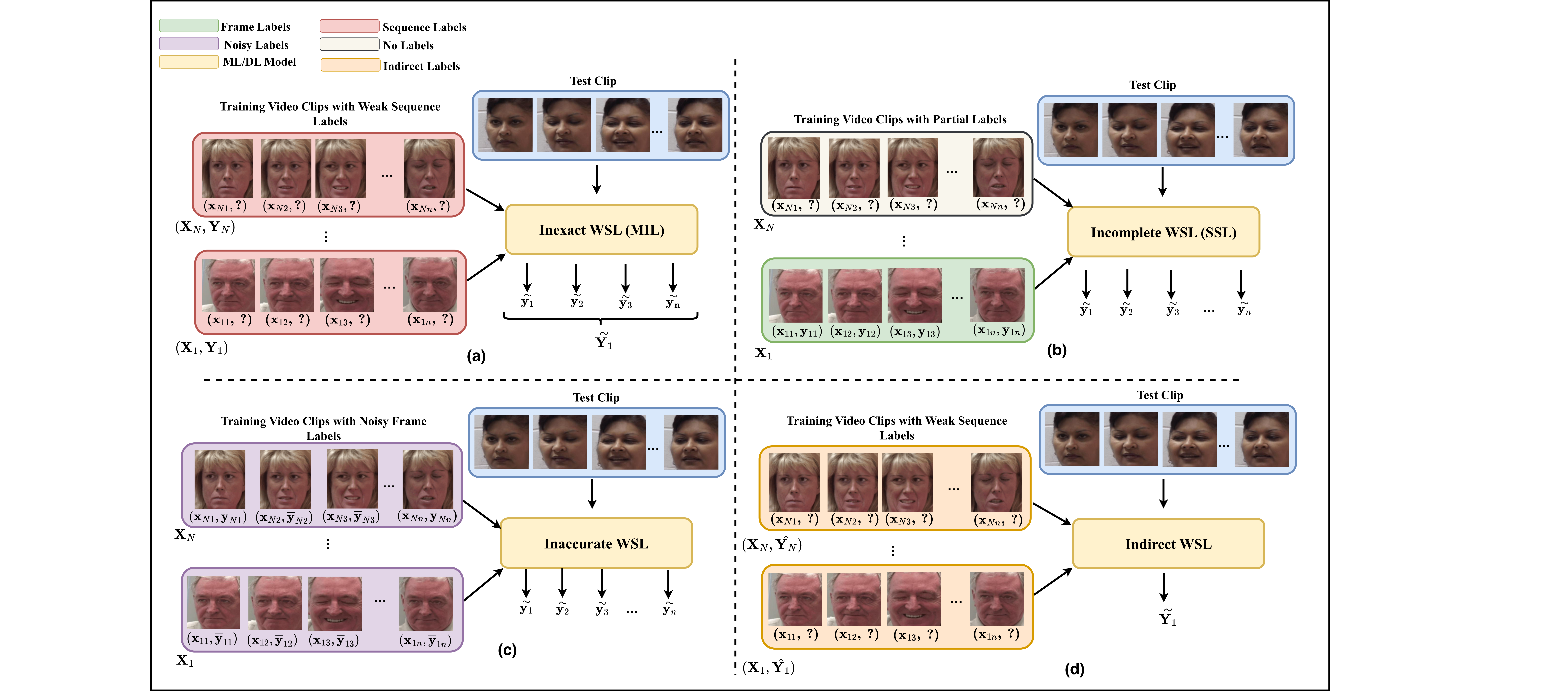}
\caption{An illustration of WSL scenarios for expression recognition in videos. (a) Inexact WSL: MIL with sequence-level expression labels. (b) Incomplete WSL: semi-supervised learning (SSL) using limited frame-level expression labels. (c) Inaccurate WSL: learning with noisy expression labels. (d) Indirect WSL: learning with indirect expression labels. ${\mathbf{X}}_1$,...,${\mathbf{X}}_N$ represents $N$ input video sequences. ${\mathbf{Y}}_1$,...,${\mathbf{Y}}_N$ represents the direct sequence-level labels, while $\hat{{\mathbf{Y}}_1}$,...,$\hat{{\mathbf{Y}}_N}$ represents indirect labels. $\overset{\sim}{{\mathbf Y}_{1}}$ represents the sequence-level predictions of test data. ${\mathbf x}_{11}$, ${\mathbf x}_{12}$, ..., ${\mathbf x}_{1n}$ denote $n$ frames of sequence ${\mathbf{X}}_1$, and ${\mathbf y}_{11}$, ${\mathbf y}_{12}$, ..., ${\mathbf y}_{1n}$ represents their respective frame-level expression labels. $\overset{\sim}{{\mathbf y}_{1}}$, $\overset{\sim}{{\mathbf y}_{2}}$, ....,  $\overset{\sim}{{\mathbf y}_{n}}$ denotes frame level predictions of test data. $\overline{\mathbf{y}}_{{11}}$, $\overline{\mathbf{y}}_{12}$, ..., $\overline{\mathbf{y}}_{1n}$ refers to noisy frame-level expression labels of sequence ${\mathbf{X}}_1$. "$\boldsymbol ?$" refers to video frames with no labels.}
\label{fig:WSLforFE}
\end{figure*}

\section{Taxonomy of Weak Supervision in Facial Affective Behavior Analysis} 
\label{WSL}

Traditionally, WSL scenarios are categorized as inexact, incomplete, and inaccurate supervision based on the availability of annotations \cite{WSL}. To better understand the landscape of weak supervision in FABA, we extend this taxonomy along two intersecting dimensions: \textbf{level of supervision} and \textbf{type of affective task}. This dual-axis taxonomy captures both the granularity of supervision (ranging from global video-level labels to frame-level intensity scores) and the specific affective task (e.g., expression recognition or AU intensity estimation). This structure provides several benefits: it (1) offers a stereo-view of individual works, highlighting how different weakly supervised annotation strategies affect various FABA tasks; (2) enables researchers to efficiently locate pertinent research aligned with their objectives and annotation paradigms; and (3) consolidates key challenges and corresponding methodologies, promoting a structured and task-specific comparative analysis. Specifically, we define four key categories of weak supervision scenarios: inexact (global), incomplete (sparse), inaccurate (noisy) supervision, and indirect (proxy) supervision.    

\subsection{Inexact (Global) Annotations}
Annotations in this category are provided at a coarse granularity, typically at the video or image-level labels, lacking precise frame or region-level labels. The core challenge lies in associating coarse global annotations with the relevant low-level instances that reflect the target behavior, while ignoring the irrelevant ones. To tackle this, Multiple Instance Learning (MIL) \cite{NIPS1997_82965d4e} and attention-based models are commonly employed to identify the most discriminative frames or regions. For instance, MIL treats each sequence as a “bag” of instances (frames or clips), allowing the model to infer which instances truly contribute to the global label, with performance being influenced by bag composition, data distribution, and label ambiguity \cite{CARBONNEAU}.


\noindent \textbf{Expression Classification:} 
This task can be formulated as static or dynamic FER, aiming to identify the spatial or temporal regions most indicative of expressions (see Fig 3a). The key challenge is to effectively model temporal dynamics to identify expression-relevant frames while simultaneously focusing on the spatial regions that convey the expression.


\noindent \textbf{AU Classification:}
The goal is to localize AU-specific facial regions using only image-level AU labels (see Fig 4a). The challenge lies in modeling dynamic spatial variations of AUs and leveraging inter-AU relationships to resolve label ambiguity and enable precise regional attribution.


\noindent \textbf{Expression or AU Intensity Estimation:}
This task involves capturing fine-grained temporal dynamics of expressions or AUs to discriminate between varying intensity levels from high-level ordinal annotations (e.g., average or maximum intensity level of instance-level outputs) \cite{MIR}.


\subsection{Incomplete (Sparse) Annotations}
Sparse supervision arises when annotations are available for only a small subset of samples, such as selected frames in a video or a limited number of AUs, leaving most data unlabeled. The central challenge is to leverage this sparse supervision to guide learning from abundant unlabeled data without amplifying errors introduced by incorrect inductive assumptions or noisy pseudo-labels. Models trained under partial annotation are also prone to overfitting observed categories while ignoring unannotated ones, which can result in biased representations.

\noindent \textbf{Expression Classification:} This scenario includes (1) sparsely annotated images or video frames, requiring generalization from labeled to unlabeled data, and (2) fully labeled source datasets combined with unlabeled target data (see Fig. 3b), requiring adaptation across domains. 
Semi-supervised methods such as pseudo-labeling and consistency regularization exploit the structure of unlabeled data to recover missing annotations \cite{Chapelle2010}. However, pseudo-labeling can amplify errors when generated labels are unreliable. Consistency regularization may encourage representation collapse, i.e., diverse inputs may have similar embeddings, leading to uninformative features. 

\begin{figure}[t!]
\centering
\includegraphics[width=0.5\textwidth]{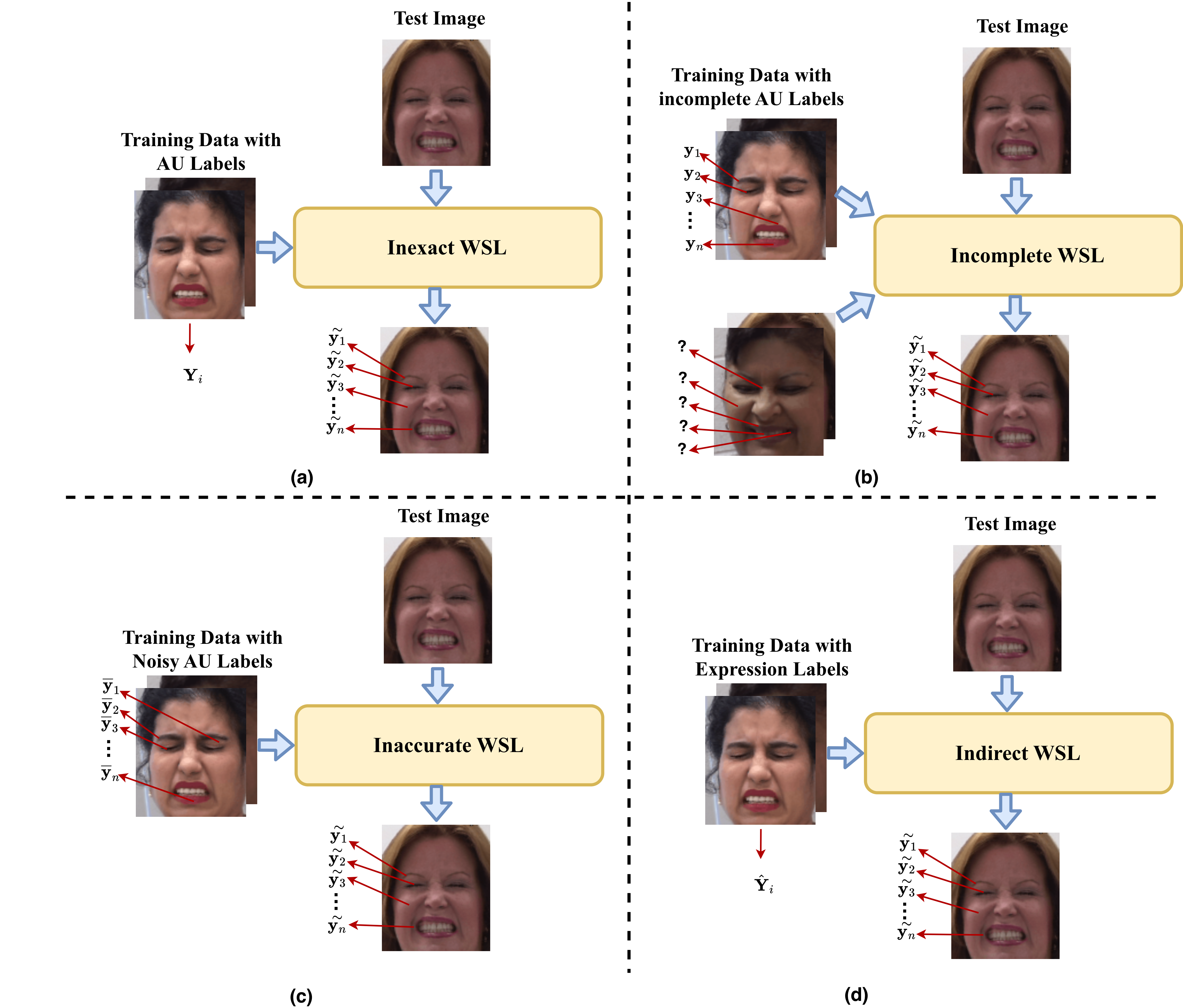}
\caption{An illustration of WSL scenarios for AU recognition in images. (a) Inexact WSL: MIL with image-level annotations. ${\mathbf Y}_{i}$ represents the image-level expression label for image $i$ (b) Incomplete WSL: SSL with limited AU annotations, (c) Inaccurate WSL with noisy AU annotations. ${\mathbf y}_1$, ${\mathbf y}_2$, ..., ${\mathbf y}_n$ denote the AU labels. $\overset{\sim}{{\mathbf y}_{1}}$, $\overset{\sim}{{\mathbf y}_2}$, ...,  $\overset{\sim}{{\mathbf y}_n}$ represent the AU predictions of test data, and $\overline{\mathbf{y}}_1$, $\overline{\mathbf{y}}_2$, ..., $\overline{\mathbf{y}}_n$ denote the noisy AU labels. (d) Indirect WSL: WSL from implicit expression labels ${\hat{\mathbf Y}_{i}}$ represents the image-level expression label for image $i$. Finally, "$\textbf{?}$" refers to the case with no annotations.} 
\label{fig:WSLforAU}
\end{figure}

\noindent \textbf{AU Classification:} The problem can be formulated as (1) limited annotations, where only a few samples have complete AU labels (see Fig. 4b), or (2) partial annotations, where only some AUs are labeled per image (missing-label setting). Challenges include modeling context-aware AU relationships for reliable label propagation and maintaining flexibility to accommodate subject-specific dynamics of different AUs.

\noindent \textbf{Expression or AU Intensity Estimation:}
This involves estimating intensity trajectories when only sparse annotations (e.g., peak or key frames) are available. Models must generalize intensity predictions across frames using temporal coherence and pseudo-label propagation.


\subsection{Inaccurate (Noisy) Annotations}
Noisy supervision requires learning robust representations that capture genuine affective cues while suppressing spurious labels introduced by subjective or ambiguous interpretation.

\noindent \textbf{Expression Classification:} 
Label noise in FER frequently arises from semantic ambiguity, where similar facial expressions may correspond to different emotions (e.g., surprise versus fear), leading to inconsistent annotations (see Fig. 3c). This ambiguity complicates the separation of overlapping affective cues from noisy supervision. Approaches such as confidence-based sample reweighting and label distribution learning mitigate these effects by modeling uncertainty rather than enforcing hard labels. Key challenges remain in obtaining reliable confidence estimates, learning uncertainty-aware representations, and handling class-specific ambiguity.




\noindent \textbf{AU Classification:} 
For AU classification, inaccurate supervision primarily stems from mislabeling of AU presence, annotator disagreement on subtle or short-lived activations, and systematic class imbalance, where less frequent AUs are under-annotated (see Fig. 4c). Such noise biases decision boundaries and weakens the learning of inter-AU dependencies, leading to implausible co-activations or missed exclusivity constraints. As a result, models trained under noisy AU supervision may overemphasize dominant or visually salient AUs while failing to reliably detect subtle or infrequent ones. This highlights the need for uncertainty-aware learning and structured relational constraints.

\noindent \textbf{Expression or AU Intensity Estimation:} 
This task introduces the additional challenges of subjective ratings, temporal imprecision in onset–apex–offset annotation, and ambiguity at low intensity levels. These factors violate smooth temporal progression and ordinal consistency, making direct regression unstable. 



\subsection{Indirect (Proxy) Annotations}
Proxy supervision refers to learning from related but indirect signals when explicit task labels are unavailable. These proxy cues often include expressions for AU recognition or textual descriptions for emotion recognition, leveraging cross-modal or inter-task correlations to transfer supervision.

\noindent \textbf{Expression Classification:}
In FER, indirect supervision often utilizes dialogue transcripts or linguistic descriptions as auxiliary sources (see Fig. 3d). The major challenges are modeling weak cross-modal correlations, handling noisy proxy labels, ensuring robust alignment, and addressing domain variability between text and vision.


\noindent \textbf{AU Classification:}
Facial expressions are widely used as proxy signals to guide AU detection (see Fig. 4d). This paradigm leverages the inherent correlations between expressions and localized AU activations. 
The main challenges include modeling inter-AU dependencies (co-occurrence and mutual exclusivity), capturing nuanced expression–AU relationships, and accounting for temporal dynamics to predict temporally coherent AUs.


\noindent \textbf{Expression or AU Intensity Estimation:}
Estimating intensity from indirect cues such as expressions or linguistic descriptions adds another layer of complexity. The challenges include learning indirect correlations between modalities, managing semantic inconsistencies, preserving fine-grained affective nuances, and maintaining temporal coherence across sequences.




\section{Methods for Classification} 
\label{WSL for classification}



\subsection{Inexact Annotations} \label{FER with inexact}
\subsubsection{Expression Recognition} \label{ED with inexact} 
Expression recognition under inexact supervision can be formulated either as spatial localization of expressions from image-level labels or as temporal localization from video-level labels, where the model must infer which spatial regions or frames actually display the target expression. Most of the existing approaches for spatial localization from image-level labels often relied on auxiliary signals (facial landmarks \cite{9956496,8974606}, facial attributes \cite{ijcai2020p145}, or AUs \cite{10582016}), which fall outside the framework of inexact supervision. We therefore focus on temporal localization in videos from video-level labels. 

\noindent \textbf{Instance Level Approaches:}
Early works on temporal expression localization cast the problem as an instance-selection task. Each video is treated as a bag of short segments (instances), and a positive video is assumed to have at least one video segment, exhibiting the expression label. Therefore, models assign a score to each segment and aggregate these scores to predict the video label. Sikka et al. \cite{SIKKA2014659} adopted this formulation using a boosting-based classifier \cite{multiple-instance-boosting-for-object-detection}, where segment scores are combined via max pooling, so that the most salient segment dominates the prediction. This design is effective when expressions are sparse and intense, and was shown to effectively capture short, high-intensity pain episodes better than treating each frame independently. Wu et al. \cite{7163116} extended this framework by coupling segment-level scores with a Hidden Markov Model \cite{1165342}, which constrains predictions to follow a plausible temporal progression (e.g., neutral–onset–apex–offset) rather than fluctuating independently across segments. By enforcing temporal consistency, this extension improves robustness to noisy or irrelevant frames. Despite these advantages, instance-centric MIL formulations are inherently biased toward peak expressions. They perform well when a single brief expression dominates the label, but tend to underperform when multiple expression cues are distributed over time, as weaker yet persistent segments may be suppressed by max-based aggregation.\\
\noindent \textbf{Bag Level Approaches:}
To better handle co-occurring or distributed expression cues, subsequent works shifted from selecting one key segment to learning richer bag representations that capture multiple affective cues and their evolution in time. Ruiz et al. \cite{Adria} proposed a multi-concept approach where each video is scored by several expression-specific classifiers, rather than forcing a single expression to explain the entire video. These expression-specific classifiers allow the model to recognize multiple emotions co-occurring in the same video. This improves multi-label accuracy compared to single-expression models, but fails to capture the temporal progression of emotions across time (neutral → onset → peak → offset). Sikka et al. \cite{7780971} addressed this problem by introducing temporal ordering constraints: instead of treating segments independently, it learns "templates" that enforce realistic progression of expressions by leveraging the gradual intensity changes and phase transitions of emotions (rather than just peak moments). This leads to better performance, but requires more complex training to learn these structured templates. Huang et al. \cite{7308029} observed that universal bag representations often fail to generalize across individuals, as individuals express emotions differently. Their personalized approach clusters video segments according to subject-specific expression patterns, then learns customized bag-level representations for each individual. This strategy improves performance on datasets with high inter-subject variability but risks overfitting when per-subject data is limited. Finally, Wang et al. \cite{10204168} noted that expressions show strong short-term correlations (nearby frames are similar) but also need long-term context to understand full dynamics. They replaced hand-crafted pooling rules with a hierarchical encoder: 3D CNNs \cite{Tran} capture motion in short clips, while LSTMs/Transformers model dependencies across the full sequence. This end-to-end design better handles subtle, distributed expressions spread across long videos, but demands more computation and can be sensitive to noisy video-level labels.

\noindent \textbf{Discussion:}
Overall, FER under inexact supervision has evolved from peak-evidence MIL, which explains the video label with a single expression \cite{SIKKA2014659,7163116} to bag-level and temporally structured models that can integrate evidence across time and accommodate multiple, co-occurring expressions \cite{Adria,7780971,10204168}. Methodologically, performance gains are largely driven by two complementary ideas. First, explicit temporal modeling, typically implemented through hierarchical short or long-term modeling \cite{10204168}, enables the model to capture both local motion patterns and long-term context, making it more effective when expression cues are weak but persistent over multiple segments \cite{7780971}. Second, progression constraints, such as ordinal or phase-based modeling, impose an interpretable ordering on expression intensity or temporal stages (e.g., onset–apex–offset), helping disambiguate gradual transitions from noise \cite{7780971}. Empirically, these design choices consistently improve robustness when expressions are temporally distributed rather than concentrated in a single peak. In practice, instance-centric MIL remains attractive when labels are reliable, and expression events are sparse and well-localized, whereas bag-level and hierarchical approaches \cite{10.5555/3295222.3295349} are better suited to complex affective scenarios involving co-occurrence, subject variability, and gradual temporal evolution. 

\subsubsection{AU Recognition} \label{AU with inexact} 

To reduce dependency on dense spatial annotations or predefined AU regions, recent WSL approaches seek to jointly infer where AUs occur and how they relate, using attention mechanisms and relational reasoning. The core idea is to let the model discover informative facial regions from image-level labels while simultaneously modeling inter-AU dependencies, rather than relying on fixed regions or external priors. Shao et al. \cite{8880512} introduced attention and relation learning (ARL), a dual-branch architecture in which one branch learns region-level attention to localize discriminative facial areas, while the other branch explicitly models relationships among AUs to capture common co-occurrence and relational patterns. This synergy allows spatial localization and semantic dependency modeling to reinforce each other under weak supervision. However, focusing solely on spatially diverse regions may fail to capture AU-specific characteristics when multiple AUs share overlapping muscle activations, often leading to suboptimal performance. To address this problem, Li et al. \cite{10154146} shifted the focus from where the model attends to how it encodes AU-specific patterns within shared spatial regions. They introduced a Self-diversified Multichannel Attention Network (SMA-Net) by employing a diversity loss that maximizes the cosine distance between attention maps of different heads, encouraging them to extract complementary features even when operating within the same region of interest. This mechanism reduces representational redundancy and mitigates attention collapse, where multiple attention heads converge on dominant AUs without capturing AU-specific features. 

\noindent \textbf{Discussion:}
Collectively, these approaches reflect a broader shift in weakly supervised AU recognition away from fixed spatial priors and toward flexible, data-driven attention mechanisms. Early works such as ARL \cite{8880512} laid the groundwork by explicitly coupling spatial attention with inter-AU relational reasoning. This dual-branch architecture 
performs well when AU interactions follow consistent relational patterns across individuals. However, in scenarios involving overlapping muscle activations or subtle AUs, representation redundancy, and attention collapse can hinder performance. Addressing this, more recent methods like SMA-Net \cite{10154146} introduce internal attention diversification to improve feature discrimination within shared spatial regions. By enforcing disagreement among attention heads, the model learns complementary representations without requiring additional labels. Together, these strategies signal a broader trend toward context-aware and representation-diverse modeling, improving robustness in AU recognition under inexact supervision.

\subsection{Incomplete Annotations} 
\label{Incomplete Annotations for classification}

\subsubsection{Expression Recognition} 
\label{IncompleteExpressionDetection} 
In this category, three dominant strategies have emerged to exploit unlabeled or sparsely labeled data: consistency regularization \cite{laine2017temporal}, pseudo-labeling \cite{Lee2013PseudoLabelT}, and hybrid frameworks that integrate both.

\noindent \textbf{Consistency Regularization (CR):} 
These strategies avoid explicit pseudo-label generation and instead encourage the model to produce stable predictions under input perturbations or alternative views of the same data. The core idea is that reliable semantic structure should be invariant to noise, or augmentation, allowing unlabeled data to contribute supervision without committing to potentially incorrect hard labels. Progressive Teacher (PT) \cite{9629313} adopted a teacher–student paradigm, where two networks initialized differently exchange supervision only on confident, low-loss samples. By progressively refining the teacher and restricting supervision to stable predictions, this strategy improves robustness while reducing the risk of confirmation bias and preserving representation diversity. Cho et al. \cite{10484312} integrate contrastive learning, transferring relational information—such as similarity and dissimilarity between samples—from labeled to unlabeled data. By selecting reliable positive and negative pairs based on learned attention patterns, they implicitly refine latent labels and promote more discriminative feature representations. In practice, consistency-based approaches are particularly effective when pseudo-labels are unreliable, but they require careful regularization to prevent representation collapse or over-smoothing when invariance constraints are enforced too aggressively.

\noindent \textbf{Pseudo-Labeling:}
These strategies exploit high-confidence model predictions as surrogate supervision for unlabeled data, but their effectiveness is often limited by label noise, spurious attention to non-discriminative regions, and severe class imbalance. To tackle the problem of attending non-discriminatory regions, Contrastive Feature Refinement Network (CFRN) \cite{10219749} explicitly refined class activation maps using a feature dropout module, which is used in a complementary fashion along with a feature emphasis module. In the feature dropout module, spurious attended regions (e.g., background or identity-specific cues) are identified using a threshold-based mask, while the feature emphasis module is used to strengthen the true key regions. By anchoring the predicted labels to more stable and meaningful visual regions, this approach helps prevent mistakes from compounding as the model continues to learn from its own predictions.
Another line of approach focuses on identifying unreliable pseudo-labels through prediction disagreement. Regularized Mixture of Predictions (ReMoP) \cite{ijcai2025p154} compared the outputs of two classifiers that learn in different ways—a center-loss–trained model \cite{Wen}, which encourages tightly grouped class features, and a simple linear classifier. When the two models disagree on a prediction, the sample is flagged as uncertain and either down-weighted or discarded. This helps the model avoid learning from noisy labels and reduces class imbalance by preventing dominant classes from dominating the training process.
 Complementarily, Dynamic Direction Learning (DDL) \cite{10955730} adopted a teacher–student framework in which pseudo-labels are generated by a teacher model calibrated using a small, balanced validation set. Rather than enforcing strict label matching, DDL aligns the feature distributions of teacher and student using Maximum Mean Discrepancy, encouraging the student to follow the teacher’s learning direction while remaining robust to noisy supervision. In practice, such distribution-level regularization stabilizes training, but introduces additional complexity and sensitivity to the quality of the validation set.

\noindent \textbf{Hybrid Approaches:}
These methods combine the strengths of pseudo-labeling and consistency-based learning by treating unlabeled samples differently based on how reliable their predictions appear. The key idea is to use confident predictions as supervision signals, while handling uncertain ones more cautiously guiding the model through training objectives that promote stability and separation between classes, without enforcing potentially incorrect labels. MarginMix \cite{Florea_eccv2020} exemplified this strategy by stabilizing pseudo-labels through both margin-based class separation and prediction aggregation. By extending the center-loss objective \cite{Wen}, MarginMix enforces large inter-class margins in the feature space, making pseudo-labels less sensitive to small perturbations. At the same time, predictions are averaged across multiple data augmentations, so only samples with consistently stable outputs are reinforced. This combination improves pseudo-label reliability while preserving clear class boundaries in the learned representations. Building on this idea, AdaCM \cite{9880204} addressed class imbalance and varying class difficulty by introducing class-specific confidence margins. Instead of using a fixed threshold for all classes, it applies higher confidence thresholds to dominant or ambiguous classes, where incorrect predictions are more likely and lower thresholds to rare or hard-to-classify classes. Confident samples are used for supervision, while uncertain ones guide representation learning through contrastive learning, improving robustness without forcing the model to rely on unreliable labels.
In practice, such hybrid designs offer more stable training under weak supervision by balancing confident learning with cautious use of uncertain predictions.

\noindent \textbf{Discussion:}
Early semi-supervised approaches largely treated high-confidence model predictions as surrogate labels for unlabeled data. While effective when decision boundaries were well separated, this strategy proved fragile in affective recognition settings, where dataset bias, class imbalance, and semantic ambiguity frequently cause confident but incorrect predictions. As a result, performance often degraded as errors accumulated over iterations. To overcome these challenges, 
recent pseudo labeling methods have moved beyond simple confidence thresholding toward strategies that explicitly improve pseudo-label reliability. These approaches refine where supervision is drawn from (e.g., focusing on spatially reliable regions \cite{10219749}), when pseudo-labels should be trusted (e.g., discrepancy-based filtering \cite{ijcai2025p154}), and how predictions are calibrated across domains or classes (e.g., distribution alignment \cite{10955730}). 
In contrast to pseudo-labeling, CR methods \cite{9629313,10484312} encourage stable predictions across augmented or perturbed versions of the same input by maintaining coherence in how inputs are represented and classified under different conditions. This makes CR especially effective when temporal structure, cross-sample relationships, or augmentation-invariant features are available. However, their success often depends on the quality of augmentations and the reliability of the consistency signals they enforce.
By leveraging the benefits of both pseudo labeling and CR strategies, 
hybrid frameworks \cite{Florea_eccv2020,9880204} enable more stable optimization and better use of unlabeled data under severe label sparsity. Overall, the field is moving away from deterministic label assignment toward uncertainty-aware representation learning, where confidence estimation governs how supervision is applied. Across methods, performance gains increasingly stem from calibrated supervision, adaptive regularization, and improved pseudo-labels. 

\subsubsection{AU Recognition} \label{Incomplete AU detection}
The central challenge in AU recognition with incomplete supervision is to exploit partially available annotations while modeling inter-AU dependencies that guide reliable label propagation.

\noindent \textbf{Multi-label Learning with Missing Labels:} 
Early approaches to AU recognition under incomplete supervision relied on structured priors and hand-crafted representations to compensate for missing labels. A common strategy was to explicitly encode prior knowledge about AU structure and co-occurrence to guide predictions when annotations were sparse. For instance, Song et al. \cite{7163081} formulated AU recognition based on the assumption that only a small subset of AUs is active in each frame, and related AUs tend to co-occur. Their Bayesian Graphical Compressed Sensing (BGCS) framework encodes known AU co-occurrence patterns and encourages groups of related AUs to activate together, enabling effective prediction even when supervision is limited. 
Complementary efforts focused on propagating known labels across related samples. Wu et al. \cite{WU20152279} introduced a multi-label learning framework with missing labels (MLML) that encourages consistency between predicted and observed labels and enforces smoothness across visually or semantically similar instances. This formulation assumes that samples, which are closer in feature space, should share similar AU activations, allowing partial annotations to influence unlabeled examples. 
Despite their effectiveness under structured assumptions, these methods were fundamentally limited by their reliance on hand-crafted features, which made it difficult to disentangle expression-specific facial movements from identity-related variations. To alleviate this limitation, Wu et al. \cite{8237688} proposed a deep AU recognition framework that integrates Restricted Boltzmann Machines (RBM) \cite{10.1145/1273496.1273596} to learn latent representations capturing AU co-occurrence and mutual exclusivity directly from data. By combining unsupervised pretraining with supervised fine-tuning on partially labeled samples, this approach marked an early transition from hand-crafted features to data-driven representation learning, while still retaining explicit modeling of AU relationships.
\noindent \textbf{Incomplete Supervision with Partially Labeled Samples:}
Recent works have increasingly adopted end-to-end learning frameworks that integrate relational reasoning, contrastive objectives, and temporal cues, with the goal of learning AU representations that are both discriminative and robust to partial supervision. A central theme across these approaches is to replace fixed structural priors with learned dependency modeling, allowing relationships between AUs to be inferred directly from data. Multi-Label Co-Regularization (MLCR) \cite{NIPS2019_8377} exemplified this shift by combining co-training with relational learning. Two networks are encouraged to produce consistent predictions on partially labeled data, using agreement as a regularizing signal, while graph convolutional networks are employed to learn AU dependencies dynamically rather than relying on predefined co-occurrence rules. This co-regularization stabilizes learning under sparse supervision, but introduces additional model complexity and sensitivity to graph construction. Complementary approaches leverage contrastive learning to compensate for partial supervision by directly learning discriminative features to promote class separation and improve representations.
Liu et al. \cite{10095708} improved performance 
by enforcing discrimination at both the instance (sample) level and the prototype (class) level, encouraging samples that share known AUs to cluster in the feature space while separating samples with conflicting annotations. To ensure that contrastive learning focuses on meaningful facial evidence, anatomical priors are used to guide attention toward plausible AU-related regions. This design helps to reduce the influence of spurious cues under noisy supervision. However, because attention masks are fixed and AU relationships are not modeled explicitly, the approach may generalize poorly across subjects, poses, or expression variations that deviate from the assumed spatial patterns. 
The Knowledge-Spreader (KS) framework \cite{Li_2023_ICCV} adopted a teacher–student architecture in which a transformer learns AU dependencies from sparsely annotated keyframes and propagates this knowledge to unlabeled frames via pseudo-labeling and temporal modeling. By unifying spatial attention, relational reasoning, and temporal consistency within a single framework, KS achieves strong performance under partial supervision, although its effectiveness depends on the representation quality of the annotated keyframes. Finally, Yan et al. \cite{9736614} integrate AU recognition with auxiliary supervisory signals, combining region-of-interest inpainting to enforce localized AU representations, optical flow to inject motion cues, and a transformer that encodes AU relationships. When coupled with MixMatch \cite{48557}, this multi-signal supervision yields robust features despite sparse labels, but at the cost of increased computational complexity and reliance on the accuracy of inpainting and motion estimation.

\noindent \textbf{Discussion:}
Approaches like BGCS \cite{7163081} and MLML \cite{WU20152279} compensate for missing AU labels by imposing static structural priors such as sparsity, co-occurrence, or label smoothness on hand-crafted features. While effective under controlled conditions, these assumptions limit their ability to adapt to complex, context-dependent AU interactions and to generalize across subjects, poses, and uncontrolled environments. 
DL–based methods \cite{8237688} alleviate these limitations by jointly learning feature representations and label structure from data, enabling flexible modeling of AU relationships. This shift paved the way for adaptive relation modeling using graph neural networks and transformers \cite{NIPS2019_8377,Li_2023_ICCV}, which infer AU co-occurrence and mutual exclusivity patterns directly from observations rather than relying on predefined rules. By updating relational structure during training, these models can better capture dataset-specific dependencies, albeit at the cost of increased computational complexity and sensitivity to noisy supervision. To further exploit partially labeled data, several methods introduce auxiliary learning signals that do not rely solely on explicit labels. Co-training and contrastive objectives \cite{NIPS2019_8377,10095708} extract supervisory signal by encouraging consistency between models or by shaping the feature space so that samples sharing known AUs are closer than those with conflicting annotations. These strategies are particularly effective when anatomical priors or reliable similarity relationships are available, but their performance can degrade when such priors are inaccurate or incomplete. Temporal teacher–student frameworks \cite{Li_2023_ICCV} 
learns spatial and relational AU representations from a small set of reliable key-frames and 
enables efficient use of sparse annotations, but its success is influenced by the representation quality of keyframes and the calibration of confidence thresholds. The field has transitioned from rigid priors and hand-crafted features toward adaptive, end-to-end frameworks that learn inter-AU relations from data, propagate supervision from partial or missing labels, and exploit temporal context to stabilize learning. 
\subsection{Inaccurate Annotations} \label{FER with inaccurate}
\subsubsection{Expression Recognition} \label{Inaccurate Expression detection}
Though expression recognition with noisy annotations has also been addressed due to pose and occlusion challenges, we focus specifically on methods that explicitly model and mitigate noisy labels.

\noindent \textbf{Label Refinement from Clean Sample Selection} 
These approaches refine supervision by explicitly separating reliable samples from uncertain ones, enabling the model to learn from cleaner signals while reducing the impact of noisy labels. Wang et al. \cite{9157210} proposed Self-Cure Network (SCN), which estimates sample reliability within each mini-batch using a self-attention–based confidence score. The key intuition is that samples consistently receiving strong, focused attention are more likely to be correctly labeled. SCN therefore prioritizes these high-confidence samples for supervision, while low-confidence samples are softly relabeled using prediction agreement, preventing early overfitting to noise. Zhang et al. \cite{zhang2021relative} introduced Relative Uncertainty Learning (RUL), which measures uncertainty through relative sample difficulty rather than relying on raw loss magnitudes or static confidence thresholds. By applying feature-level mixup, RUL exposes inconsistencies in facial representations and down-weights unreliable features and labels. This design allows the model to retain informative but hard samples, instead of discarding them as noise. Building on this idea, Nie et al. \cite{10614222} proposed the IntraClass-Divide framework, which models loss distributions separately for each expression class using Gaussian Mixture Models (GMMs). Unlike global GMM approaches that often confuse inherently difficult expressions with mislabeled samples, this class-specific modeling better captures heterogeneous noise patterns and enables more precise separation of clean and noisy data. Together, these methods demonstrate how uncertainty estimation and class-aware sample selection can stabilize learning under noisy annotations without requiring additional supervision.

\noindent \textbf{Label Distribution Learning (LDL):} 
LDL addresses annotation noise and ambiguity by replacing hard labels with soft label distributions \cite{7439855} that encode uncertainty over multiple affective states. This formulation is particularly well suited to facial expression analysis, where annotators often disagree or where expressions lie between discrete categories. Early LDL-based approaches focused on estimating reliable label distributions from noisy supervision. She et al. \cite{9578636} inferred soft labels by aggregating predictions from multiple auxiliary classifiers and weighting them using uncertainty estimates. By encouraging samples with similar feature representations to share similar label distributions, the method smoothens inconsistent annotations while preserving inter-class separability in the feature space. Subsequent works incorporated teacher-guided refinement to stabilize distribution estimation. Lukov et al. \cite{10.1007/978-3-031-19775-8_38} adopted a mean-teacher framework, where averaged teacher predictions are used to generate soft label distributions for the student network. High-confidence logits are converted into soft probability distributions, while low-confidence classes receive small but non-zero probabilities. This discourages the model from making overly confident predictions on uncertain inputs. To further reduce the impact of noisy or unreliable outputs, the process is regularized using uncertainty estimation techniques \cite{9157210}.
More recent methods addressed the dynamic nature of label noise during training. Self-Paced Label Distribution Learning (SPLDL) \cite{10.1145/3503161.3547960} introduced a curriculum strategy that progressively learns hard and noisy samples. Using a predefined label distribution generator \cite{Zhao_Liu_Zhou_2021}, the model first learns from low-loss, reliable instances and gradually adapts the label distributions as training progresses, improving robustness to severe and class-dependent noise. Overall, LDL has evolved from static, heuristically defined soft labels to adaptive, uncertainty-aware, and curriculum-driven distributions. This progression improves robustness to ambiguous supervision but introduces additional complexity and sensitivity to distribution initialization and uncertainty estimation quality.

\noindent \textbf{Ensembling for Noise Mitigation:} 
Multi-view aggregation methods mitigate annotation noise by combining predictions or feature representations obtained from multiple annotations, model instances, or data perturbations, thereby reducing the influence of any single unreliable view. 
Zeng et al. \cite{10.1007/978-3-030-01261-8_14} proposed a framework that explicitly treats noisy labels as multiple inconsistent observations of an underlying latent ground truth. Their method trains models on these conflicting annotations and then infers latent labels by aggregating pseudo-annotations through log-likelihood maximization, effectively filtering noise by favoring label configurations that are globally consistent with the data. Subsequent works shifted attention from label aggregation to feature- and attention-level consistency. Zhang et al. \cite{10.1007/978-3-031-19809-0_24} observed that, under noisy supervision, deep models tend to overfit spurious local regions that correlate with incorrect labels. To counter this, they enforce consistency between attention maps generated from augmented versions of the same image, encouraging the model to rely on stable, globally informative facial regions rather than noisy cues. More recently, Zheng et al. \cite{Zheng_Li_Zhang_Wu_Cao_Ding_2023} proposed a geometry-aware framework that improves robustness to label noise and class imbalance. Instead of treating all samples equally, the method detects mislabeled data by measuring how sensitive a sample’s features are to small adversarial changes in a geometry-informed space. Samples that react abnormally are flagged as likely noisy. The model then applies a divide-and-conquer strategy, treating clean and noisy data separately and rebalancing training to avoid bias toward dominant classes. By explicitly incorporating facial geometric structure, the approach provides more reliable training even in the presence of annotation errors. Overall, these approaches reflect a progression from label-level voting schemes toward representation- and geometry-aware consistency mechanisms, with increasing emphasis on identifying unreliable supervision and enforcing stability across views to guide learning under noisy annotations.

\noindent \textbf{Discussion:}  
Clean-sample selection methods have evolved from global, loss-based heuristics to class-aware and uncertainty-weighted filtering, explicitly accounting for the fact that annotation noise often varies across expressions \cite{9157210,zhang2021relative,10614222}. By estimating sample reliability in a class-conditional manner and weighting supervision accordingly, these approaches improve robustness to class-specific noise patterns. However, their effectiveness depends critically on accurate confidence estimation: when confidence scores are poorly calibrated, clean and noisy samples become difficult to separate, limiting their benefit. LDL addresses noise from a different perspective by replacing hard labels with soft distributions that encode ambiguity and annotator disagreement. Early LDL approaches relied on auxiliary prediction heads to estimate these distributions \cite{9578636}, while more recent methods incorporate teacher-guided smoothing and self-paced curriculum that progressively refine soft labels as training stabilizes \cite{10.1007/978-3-031-19775-8_38,10.1145/3503161.3547960}. This formulation is particularly effective when expressions are inherently ambiguous or subjectively annotated, but its performance hinges on well-calibrated soft labels and reliable teacher signals. Ensembling and consistency-based methods take yet another route, stabilizing learning by enforcing agreement across multiple views, model instances, or feature geometries \cite{10.1007/978-3-030-01261-8_14,10.1007/978-3-031-19809-0_24,Zheng_Li_Zhang_Wu_Cao_Ding_2023}. Rather than correcting labels explicitly, these approaches constrain representations to remain invariant under perturbations, which is especially effective when noise manifests as spurious saliency or fragmented supervision. 
Across these methods, a unifying trend is a shift away from deterministic relabeling toward probabilistic, representation-centric noise handling, emphasizing uncertainty awareness, agreement-based regularization, and calibration. Persistent challenges remain, including structured label noise, underrepresented expressions, and biases correlated with identity or context. These limitations increasingly motivate hybrid pipelines that combine soft-label modeling with consistency constraints, and confidence calibration with class- or geometry-aware sample selection.

\subsubsection{AU Recognition} \label{Inaccurate AU detection}
Recent approaches mitigate label noise in weakly supervised AU recognition by exploiting inherent data structure, using representation refinement and consistency constraints instead of explicit label correction. 
Zhao et al. \cite{Zhao_2018_CVPR} addressed AU learning by collecting large-scale web images through expression-based keyword searches, treating the retrieved expression labels as noisy AU annotations. 
Their method employs a scalable clustering to refine these weak labels and learns a spectral embedding that pulls together visually and semantically similar faces, followed by rank-order clustering to form confident AU-specific groups. These clusters are re-labeled via majority vote and used to train AU detectors. This weakly supervised pipeline effectively denoises noisy labels without manual annotation, achieving high agreement with expert labels. The approach relies on the accuracy of initial tags and models each AU independently, but shows that embedding-guided relabeling can generate high-quality supervision from large, noisy web image collections.
Fabian et al. \cite{8237690} introduced a global–local learning objective that strengthens region-level AU cues while enforcing structural consistency within images. The key intuition is that the spatial organization of facial movements remains stable even with noisy annotations. Encouraging agreement between local AU evidence and global facial structure helps prevent overfitting to annotation errors. More recently, hybrid co-training frameworks have combined clean-sample selection with noise-aware regularization. ReCoT \cite{Li_2023_BMVC} separated training into two complementary streams: a clean network trained on class-balanced, high-confidence samples, and a noisy network regularized using all available data. Cross-view consistency aligns predictions between these networks, while temporal smoothing and fusion progressively refine noisy labels, mitigating class imbalance and reducing error accumulated over time. 

\noindent \textbf{Discussion:} 
Embedding- and clustering-based approaches \cite{Zhao_2018_CVPR} are most effective when large weakly labeled datasets exhibit coherent latent structure, providing lightweight denoising with modest computational cost, but remaining sensitive to clustering granularity and representation quality. Global–local consistency methods \cite{8237690} primarily improve spatial reliability under coarse supervision, helping suppress spurious attention but offering limited correction for systematic class-level noise. Co-training frameworks such as ReCoT \cite{Li_2023_BMVC} is found to be promising in handling severe noise and class imbalance by explicitly separating trusted from unreliable supervision and enforcing cross-view agreement, albeit with higher complexity and greater sensitivity to hyperparameter choices (e.g., confidence thresholds and agreement criteria). Overall, these approaches illustrate a shift from confidence-based filtering toward structure- and agreement-driven noise handling, emphasizing robustness through data organization, representation constraints, and training dynamics rather than direct label correction.

\subsection{Indirect (Proxy) Annotations} \label{FER with inaccurate}
Indirect supervision leverages auxiliary signals such as captions, dialogue, or language-model outputs—that are semantically related to facial expressions but do not provide explicit emotion annotations. The central intuition is that language carries rich affective information that can guide representation learning when direct labels are unavailable, allowing models to scale to large, in-the-wild datasets. 

\subsubsection{Expression Recognition} \label{AU detecion from expressions}
Recent vision–language approaches replace categorical emotion labels with language priors learned through cross-modal alignment. EmotionCLIP \cite{zhang2023learning} aligned facial features with dialogue-based sentiment using contrastive learning within a CLIP-style video–text framework \cite{pmlr-v139-radford21a}. To improve grounding in affective signals, it uses subject-aware context encoding to help the model attend to nonverbal cues (like facial expressions and posture) and employs sentiment-guided contrastive learning to align visual features with verbal emotion cues. Trained on multimodal data from films and vlogs, it can learn affect-rich representations without relying on manual emotion labels. However, since speech and facial expressions may not always align temporally or semantically (e.g., sarcasm, off-screen dialogue), the weak supervision signal can introduce noise—particularly at the frame level. 
EmoCLIP \cite{foteinopoulou_emoclip_2024} addressed this limitation by replacing coarse emotion categories with sample-level descriptive captions, which provide richer and more context-aware supervision. They fine-tuned CLIP to directly align video frames with their associated textual descriptions using contrastive objectives, leading to richer supervision and better generalization to subtle expressions. 
This benefit, however, comes at the cost of relying on high-quality captions, which often require expert annotation or careful curation. More recently, Exp-CLIP \cite{Zhao_2025_WACV} removes human annotation entirely by aligning facial representations with LLM-derived expression embeddings, enabling label-efficient and fine-grained zero-shot FER. By training a lightweight projection head to match facial features with LLM-derived descriptions through contrastive learning, the model learns expression-specific semantics without manual labels, achieving fine-grained, zero-shot FER with strong generalization across diverse datasets. However, its effectiveness depends on the quality of prompts, the semantic coverage of the generated embeddings, and the diversity of the unlabeled visual data used for alignment. 

\noindent \textbf{Discussion:}
Vision–language weak supervision for FER replaces explicit emotion labels with language priors, introducing a spectrum of trade-offs between scalability, semantic precision, and robustness. Dialogue-driven supervision \cite{zhang2023learning} scales naturally with large volumes of multimodal video, making it attractive for data-hungry models, but is prone to modality drift when speech content is weakly aligned with facial expressions, particularly in the absence of frame-level grounding. Caption-driven supervision \cite{foteinopoulou_emoclip_2024} offers more targeted and interpretable affective cues, improving semantic specificity, but its effectiveness depends on caption quality and limits throughput due to annotation or curation costs. LLM-derived embedding supervision \cite{Zhao_2025_WACV} is the most label-efficient alternative, enabling fine-grained zero-shot recognition, yet it remains sensitive to prompt design, embedding calibration, and the semantic coverage of the unlabeled corpus. Across these variants, empirical evidence suggests that temporal grounding (e.g., frame- or segment-level alignment), face-centric constraints to reduce off-screen or contextual leakage, and normalization of linguistic supervision (prompt or caption style and vocabulary) are critical for stabilizing training and preserving fine-grained facial cues. Collectively, these approaches highlight a core trade-off in indirect supervision for FER: reducing annotation cost and improving scalability comes at the expense of increased sensitivity to language priors and cross-modal alignment quality.

\subsubsection{AU Recognition} \label{AU detecion from expressions}
Motivated by the lower cost and wider availability of expression-level annotations, 
early approaches relied on fixed expression–AU mappings derived from statistical correlations (see Table \ref{Expressions - AUs}). Ruiz et al. \cite{7410779} introduced a hidden task learning framework that transfers supervision from expressions to AUs using predefined expression–AU associations. During training, the model predicts both expressions and AUs, but only expression labels are supervised directly. A consistency loss is explored ensures that the predicted AUs align with the expected AU pattern of the predicted expression, allowing gradients to train the AU branch without AU labels. 
While conceptually simple, this method assumed static and independent mappings, limiting its ability to model joint AU dependencies or adapt to individual or contextual variability.
To address these limitations, subsequent probabilistic methods introduced probabilistic priors to model uncertainty in expression–AU relationships. These methods typically generate pseudo AU labels from soft priors (e.g., expression–AU statistics) and then refine these labels using models that encode structural dependencies among AUs. 
For instance, Wang et al. \cite{8329513} used RBM to learn the joint distribution of AUs by modeling co-activation and mutual exclusivity between AUs as a latent binary representation, where visible units correspond to AU labels and hidden units capture high-level dependencies. During training, pseudo AU labels (inferred from expression priors) are input as constraints, and the RBM learns to adjust the joint AU predictions to reflect plausible co-occurrence patterns. This allows the model to correct noisy pseudo-labels by enforcing biologically and contextually realistic AU combinations. In contrast, Peng et al. \cite{8578331} explored adversarial framework by guiding the AU predictor using a discriminator that distinguishes between the predicted AU distributions and a prior distribution derived from expression–AU associations. The AU predictor is trained to "fool" the discriminator by producing AU distributions that resemble the prior, while the discriminator learns to detect mismatches. This adversarial objective ensures that AU outputs align with known priors while still allowing flexibility to deviate when needed — thus balancing guidance and adaptability.
These strategies improve flexibility over fixed mappings, but their performance remains sensitive to the quality and transferability of handcrafted priors. To tackle this problem, ordinal and fusion-based frameworks aimed to improve flexibility and robustness by moving beyond reliance on a single prior. Wang et al. \cite{8472814} incorporated domain knowledge in the form of partial ordering constraints between AUs. For example, if AU12 is typically stronger than AU14 in expressions of happiness, the model enforces this ordering during learning by introducing a ranking loss or ordinal margin. This guides the model to predict relative AU activations even when absolute labels are unavailable, helping it focus on relational structure rather than precise intensity values. On the other hand, LP-SM \cite{8578634} addressed the unreliability of individual priors by combining multiple weak cues such as expression-to-AU mappings, region-based saliency, and statistical priors—into a single model in a probabilistic fusion framework. 
By learning to weigh each prior dynamically based on the consistency of its predictions with the data, LP-SM can attenuate the influence of misleading sources while emphasizing more reliable ones. This fusion improves robustness within a dataset, though generalization across datasets remains difficult when expression–AU associations vary significantly. 
To further enhance weak supervision, dual-task and dual-direction frameworks introduce stronger inductive constraints by coupling AU recognition with face synthesis, thereby enforcing consistency between prediction and visual appearance. In WSDL \cite{8712447}, the model is trained in a two-stage setup: first, an AU recognition branch predicts AU activations from facial images, and then a face generation module (a decoder or generator network) attempts to reconstruct the original face image conditioned on the predicted AUs. Based on the reconstruction error, the self-reconstruction loop allows the model to verify the plausibility of AU predictions by checking whether they can explain the visual input. 
Peng et al. \cite{peng2019dual} extend this idea with a GAN-based framework where a generator synthesizes face images from predicted AUs, and a discriminator checks whether the output matches both the real distribution and AU configuration. In parallel, an AU recognition branch is trained on both real and generated faces, promoting consistency between AU predictions and facial appearance. This bidirectional setup strengthens learning under weak supervision by ensuring that predicted AUs are both discriminative and visually plausible.

\begin{table}[ht!]
\caption{List of AUs observed in expressions \cite{FE}.}
\label{Expressions - AUs}
\centering
 \begin{tabular}{|c | c|} 
 \hline   
 \textbf{Expression} & \textbf{AUs} \\ [0.5ex] 
 \hline\hline
    Anger & 4, 5, 7, 10, 17, 22-26   \\ [0.5ex] 
    Disgust & 9, 10, 16, 17, 25, 26 \\ [0.5ex]
    Fear & 1, 2, 4, 5, 20, 25, 26, 27 \\ [0.5ex]
    Happiness & 6, 12, 25 \\ [0.5ex]
    Sadness & 1, 4, 6, 11, 15, 17 \\ [0.5ex]
    Surprise & 1, 2, 5, 26, 27 \\ [0.5ex]
    Pain & 4, 6, 7, 9, 10, 12, 20, 25, 26, 27, 43 \\ [0.5ex]
 \hline
 \end{tabular}
\end{table}

\noindent \textbf{Discussion:}
AU recognition from expression supervision has progressed from fixed expression–AU mappings toward probabilistic and bidirectional formulations that better accommodate ambiguity, subject variability, and inter-AU dependencies. Fixed mappings \cite{7410779} are effective when expression–AU relationships are reliable, and variation is limited, but they break down under cross-subject differences and context-dependent expression realizations. Probabilistic pseudo-AU approaches \cite{7410779,8578331} fuse multiple weak cues such as expression tags, co-occurrence statistics, or region scores, and are preferable when supervision is heterogeneous; however, they remain sensitive to bias or mismatch in the underlying priors. Ordinal and unified-prior models \cite{8472814,8578634} provide a more stable alternative when absolute AU probabilities are unreliable but relative relationships between AUs remain consistent, allowing learning to proceed from partial yet robust structural information. Dual-task or dual-direction frameworks \cite{8712447,peng2019dual} impose the strongest inductive constraints by requiring predicted AUs to both explain and reconstruct facial configurations, yielding improved robustness under weak supervision at the cost of increased computational burden and tuning complexity. Across these paradigms, recurring challenges—including biased priors, cross-dataset drift, and optimization overhead—highlight the need for uncertainty-aware weighting of pseudo-labels, deformation-sensitive localization to handle non-rigid facial motion, and lightweight temporal grounding to stabilize learning without relying on dense AU annotations.

\subsection{Model Evaluation Protocols} \label{exp for classification}



\begin{table*}[!t]
\renewcommand{\arraystretch}{1.4}
\caption{Performance of state-of-the-art facial expression classification approaches under various WSL settings on the most widely evaluated datasets. 'Conventional' refers to the training and testing partitions provided by the dataset organizers. $\dagger$ denotes that the model is trained on both RAF-DB and AffectNet training sets. For incomplete supervision, AffectNet is used as unlabeled data.}
\label{Results on Classification of expressions}
\centering
\resizebox{\textwidth}{!}{
\begin{tabular}{|c|c|c|c|c|c|c|c|}
\hline
\textbf{WSL Setting} & \textbf{Dataset} &\textbf{Method} & \textbf{Task} & \textbf{Features} & \textbf{Learning Model} & \textbf{Validation} & \textbf{Accuracy}  \\ 
\hline \hline
\multirow{6}{*}{Inexact} & \multirow{6}{*}{UNBC - McMaster} & Sikka et al. \cite{SIKKA2014659} (FG 2013) & Pain [2 classes] & BoW & MILBOOST & LOSO & 83.70 \\ \cline{3-8} 
                           &   & Wu et al. \cite{7163116} (FG 2015) & Pain [2 classes] & geometric & HMM & LOSO & 85.23  \\ \cline{3-8} 
                            &   & Adria et al. \cite{Adria} (BMVC 2014)& Pain [2 classes] & 3D-SIFT & RMC-MIL & LOSO & 85.70\\ \cline{3-8}
                            &   & Sikka et al. \cite{7780971} (CVPR 2016) & Pain [2 classes] & SIFT & LOMo & LOSO & 87.00\\ \cline{3-8}
                            &   & Huang et al. \cite{7308029} (TAFFC 2016) & Pain [2 classes] & geometric & P-MIL & LOSO &  84.40\\ 
                            \hline \hline
\multirow{8}{*}{Incomplete} & \multirow{4}{*}{RAF-DB} & Li et al. \cite{9880204} (CVPR 2022) & Expression [7 classes] & ResNet-18 & Ada-CM & Conventional & 89.28 \\ \cline{3-8} 
                           &   & Jiang et al. \cite{9629313} (TAFFC 2023) & Expression [7 classes] & ResNet-18 & PT & Conventional & 88.69  \\ \cline{3-8} 
                            &   & Li et al. \cite{ijcai2025p154} (IJCAI 2025)& Expression [7 classes] & ResNet-18 & ReMoP & Conventional & 90.29\\ \cline{3-8}
                            &   & Li et al. \cite{10955730} (TAFFC 2025) & Expression [7 classes] & ResNet-18 & DDL & Conventional & 89.58\\
                            \cline{2-8}
                            & \multirow{4}{*}{FER+} & Li et al. \cite{9880204} (CVPR 2022) & Expression [8 classes] & ResNet-18 & Ada-CM & Conventional & 87.80 \\ \cline{3-8} 
                           &   & Jiang et al. \cite{9629313} (TAFFC 2023) & Expression [8 classes] & ResNet-18 & PT & Conventional & 86.60  \\ \cline{3-8} 
                            &   & Li et al. \cite{ijcai2025p154} (IJCAI 2025)& Expression [8 classes] & ResNet-18 & ReMoP & Conventional & 89.06\\ \cline{3-8}
                            &   & Li et al. \cite{10955730} (TAFFC 2025) & Expression [8 classes] & ResNet-18 & DDL & Conventional & 89.16\\ 
                            \hline \hline
\multirow{22}{*}{Inaccurate} & \multirow{8}{*}{RAF-DB}  
 & Zeng et al. \cite{10.1007/978-3-030-01261-8_14}$^\dagger$  (ECCV 2018) & Expression [7 classes] & ResNet \cite{7780459} & IPA2LT & Conventional  & 86.77\\ \cline{3-8}
& & Wang et al. \cite{9157210} (CVPR 2020) & Expression [7 classes] & ResNet-18 & SCN & Conventional & 87.03\\ \cline{3-8} 
& & She et al. \cite{9578636} (CVPR 2021) & Expression [7 classes] & ResNet-18 & DMUE &Conventional & 88.76\\ \cline{3-8}
& & Zhang et al. \cite{zhang2021relative} (NeurIPS 2021)  & Expression [7 classes] & ResNet-18 & RUL & Conventional & 88.98 \\ \cline{3-8} 
& & Lukov et al. \cite{10.1007/978-3-031-19775-8_38} (ECCV 2022) & Expression [7 classes] & ResNet-18 & SLS & Conventional & 90.42\\ \cline{3-8} 
& & Zhang et al. \cite{10.1007/978-3-031-19809-0_24} (ECCV 2022) & Expression [7 classes] & ResNet-18 & EAC & Conventional & 89.99\\ \cline{3-8} 
& & Shao et al. \cite{10.1145/3503161.3547960} (ACM MM 2022) & Expression [7 classes] & ResNet-18 & SPLDL &Conventional & 88.59\\ \cline{3-8}
& & Zheng et al. \cite{Zheng_Li_Zhang_Wu_Cao_Ding_2023} (AAAI 2023) & Expression [7 classes] & ResNet-18 & GAAVE & Conventional & 91.53\\ \cline{2-8} 
& \multirow{8}{*}{AffectNet}  
& Zeng et al. \cite{10.1007/978-3-030-01261-8_14}$^\dagger$  (ECCV 2018) & Expression [8 classes] & ResNet \cite{7780459} & IPA2LT & Conventional  & 55.71\\ \cline{3-8}
& & Wang et al. \cite{9157210} (CVPR 2020) & Expression [8 classes] & Resnet-18 & SCN & Conventional & 60.23\\ \cline{3-8} 
& & Shao et al. \cite{10.1145/3503161.3547960} (ACM MM 2022) & Expression [7 classes] & ResNet-18 & SPLDL &Conventional & 59.76\\ \cline{3-8}
& & She et al. \cite{9578636} (CVPR 2021) & Expression [8 classes] & ResNet-18 & DMUE &Conventional & 62.84\\ \cline{3-8}
& & Zhang et al. \cite{zhang2021relative} (NeurIPS 2021)  & Expression [8 classes] & ResNet-18 & RUL & Conventional & 61.43 \\ \cline{3-8} 
& & Lukov et al. \cite{10.1007/978-3-031-19775-8_38} (ECCV 2022) & Expression [8 classes] & ResNet-18 & SLS & Conventional & 62.69\\ \cline{3-8} 
& & Zhang et al. \cite{10.1007/978-3-031-19809-0_24} (ECCV 2022) & Expression [7 classes] & ResNet-18 & EAC & Conventional & 65.32 \\ \cline{3-8} 
& & Zheng et al. \cite{Zheng_Li_Zhang_Wu_Cao_Ding_2023} (AAAI 2023) & Expression [8 classes] & ResNet-18 & GAAVE & Conventional & 63.25\\ \cline{2-8}

& \multirow{6}{*}{FER+}   & Wang et al. \cite{9157210} (CVPR 2020) & Expression [8 classes] & ResNet-18 & SCN & Conventional & 89.35\\ \cline{3-8}
& & She et al. \cite{9578636} (CVPR 2021) & Expression [8 classes] & ResNet-18 & DMUE &Conventional  & 88.64\\ \cline{3-8}
& & Zhang et al. \cite{zhang2021relative} (NeurIPS 2021)  & Expression [8 classes] & ResNet-18 & RUL & Conventional & 88.75\\ \cline{3-8}
& & Zhang et al. \cite{10.1007/978-3-031-19809-0_24} (ECCV 2022) & Expression [8 classes] & ResNet-18 & EAC & Conventional & 89.64\\ \cline{3-8}
& & Lukov et al. \cite{10.1007/978-3-031-19775-8_38} (ECCV 2022) & Expression [8 classes] & ResNet-18 & SLS & Conventional & 88.60\\ 
\cline{3-8} 
& & Zheng et al. \cite{Zheng_Li_Zhang_Wu_Cao_Ding_2023} (AAAI 2023) & Expression [8 classes] & ResNet-18 & GAAVE & Conventional & 89.29\\ \hline \hline 
\multirow{3}{*}{Indirect} & \multirow{3}{*}{MAFW}  
 & Zhang et al. \cite{zhang2023learning}  (CVPR 2023) & Expression [43 classes] & ViT & EmotionCLIP & 5-fold  & 11.65\\ \cline{3-8}
 && Foteinopoulou et al. \cite{foteinopoulou_emoclip_2024}  (FG 2024) & Expression [43 classes] & ViT & EmoCLIP & 5-fold  & 33.49\\ \cline{3-8}
 && Zhang et al. \cite{Zhao_2025_WACV}  (WACV 2025) & Expression [43 classes] & ViT & ExpCLIP & 5-fold  & 26.98\\ \cline{3-8}

\hline
\end{tabular}}
\end{table*}

\begin{table*}[!t]
\renewcommand{\arraystretch}{1.4}
\caption{Performance of state-of-the-art AU classification approaches under WSL settings on most widely evaluated datasets. For incomplete supervision, EmotioNet \cite{7780969} is used as unlabeled data.}
\label{Results on Classification of action units}
\centering
\resizebox{\textwidth}{!}{
\begin{tabular}{|c|c|c|c|c|c|c|c|}
\hline
\textbf{WSL Setting} & \textbf{Dataset} &\textbf{Method} & \textbf{Task} & \textbf{Features} & \textbf{Learning Model} & \textbf{Validation} & \textbf{F1-Score}  \\ 
\hline \hline
\multirow{4}{*}{Inexact} & \multirow{2}{*}{DISFA} 
& Shao et al. \cite{8880512} (TAFFC 2022) & AU [8 AUs] & CNN & ARL & 3-fold &  58.7\\ \cline{3-8}
& & Li et al. \cite{10154146} (TAFFC 2024) & AU [8 AUs] & CNN & SMA-Net & 3-fold &  62.4\\ \cline{2-8}
& \multirow{2}{*}{BP4D}  
& Shao et al. \cite{8880512} (TAFFC 2022) & AU [12 AUs] & CNN & ARL & 3-fold &  61.1\\ \cline{3-8}
& & Li et al. \cite{10154146} (TAFFC 2024) & AU [12 AUs] & ViT & SMA-Net & 3-fold &  65.2\\ \hline \hline
\multirow{4}{*}{Incomplete} & \multirow{2}{*}{DISFA} 
& Niu et al. \cite{NIPS2019_8377} (NeurIPS 2019) & AU [8 AUs] & ResNet-34 & MLCR & 3-fold &  59.2\\ \cline{3-8}
& & Yan et al. \cite{9736614} (TMM 2023) & AU [8 AUs] & ResNet-18 & WSRTL & 3-fold &  64.6\\ \cline{2-8}
& \multirow{2}{*}{BP4D} & Niu et al. \cite{NIPS2019_8377} (NeurIPS 2019) & AU [12 AUs] & ResNet-34 & MLCR & 3-fold &  59.8\\ \cline{3-8}
& & Yan et al. \cite{9736614} (TMM 2023) & AU [12 AUs] & ResNet-18 & WSRTL & 3-fold &  65.9\\ 
\hline \hline

\multirow{16}{*}{Indirect} & \multirow{4}{*}{UNBC - McMaster} 
& Ruiz et al. \cite{7410779} (ICCV 2015) & AU [6 AUs] & SIFT & HTL & 5-fold &  23.5\\ \cline{3-8}
&   & Wang et al. \cite{8329513} (TAFFC 2020) & AU [6 AUs] & geometric & RBM-P & 5-fold &  35.1\\ \cline{3-8}
                            &   & Wang et al. \cite{8712447} (TMM 2019) & AU [6 AUs]  & geometric & WSDL & 5-fold &  40.0\\ \cline{3-8}
                            &   & Peng et al. \cite{8578331} (CVPR 2018)& AU [6 AUs]  & geometric & RAN & 5-fold & 37.6\\ \cline{2-8}
                            & \multirow{5}{*}{CK+}  & Ruiz et al. \cite{7410779} (ICCV 2015) & AU [12 AUs] & SIFT & HTL & 5-fold &  46.9\\ \cline{3-8}
                            &   & Wang et al. \cite{8329513} (TAFFC 2020) & AU [12 AUs] & geometric & RBM-P & 5-fold &  70.5\\ \cline{3-8}
                            &   & Wang et al. \cite{8712447} (TMM 2019)& AU [12 AUs] & geometric & WSDL & 5-fold &  74.0\\ \cline{3-8}
                            &   & Peng et al. \cite{8578331} (CVPR 2018)& AU [12 AUs] & geometric & RAN & 5-fold & 71.5\\ \cline{3-8}
                            &   & Zhang et al. \cite{8578634} (CVPR 2018)& AU [8 AUs] & LBP & LP-SM & 5-fold & 73.2\\ \cline{2-8}
                            & \multirow{5}{*}{MMI} & Ruiz et al. \cite{7410779} (ICCV 2015)& AU [14 AUs] & SIFT & HTL & 5-fold &  43.1\\ \cline{3-8}
                            &   & Wang et al. \cite{8329513} (TAFFC 2020) & AU [13 AUs] & geometric & RBM-P & 5-fold &  51.6\\ \cline{3-8}
                            &   & Wang et al. \cite{8712447} (TMM 2019)& AU [13 AUs] & geometric & WSDL & 5-fold &  53.0\\ \cline{3-8}
                            &   & Peng et al. \cite{8578331} (CVPR 2018)& AU [13 AUs] & geometric & RAN & 5-fold & 52.0\\ \cline{3-8}
                            &   & Zhang et al. \cite{8578634} (CVPR 2018) & AU [8 AUs] & LBP & LP-SM & 5-fold & 48.1\\ \cline{2-8}
                            & \multirow{2}{*}{DISFA} & Ruiz et al. \cite{7410779} (ICCV 2015)& AU [12 AUs] & SIFT & HTL & 5-fold &  37.1\\ \cline{3-8}
                            &   & Wang et al. \cite{8329513} (TAFFC 2020)& AU [12 AUs] & geometric & RBM-P & 5-fold &  42.4\\ \hline  
\end{tabular}}
\end{table*}


\begin{figure}[t!]
\centering
\includegraphics[width=0.39 \textwidth]{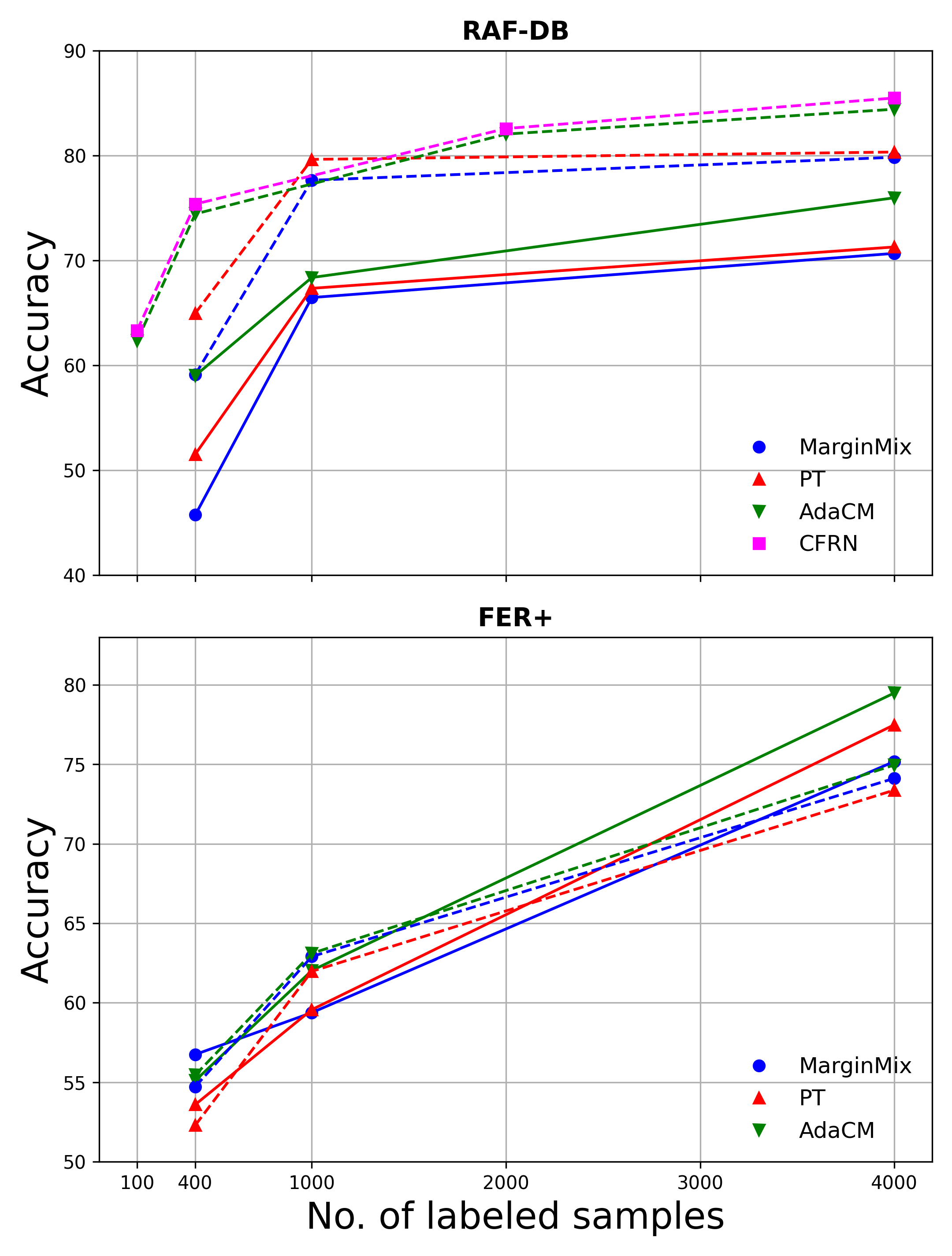}
\caption{Accuracy of expression classification methods with incomplete annotations on RAF-DB (top) and FER+ (bottom) datasets. Dotted and dashed lines denote performance using Resnet-18 and WideResnet-28-2, respectively.} 
\label{fig:SSLonExp}
\end{figure}

\begin{figure}[t!]
\centering
\includegraphics[width=0.39 \textwidth]{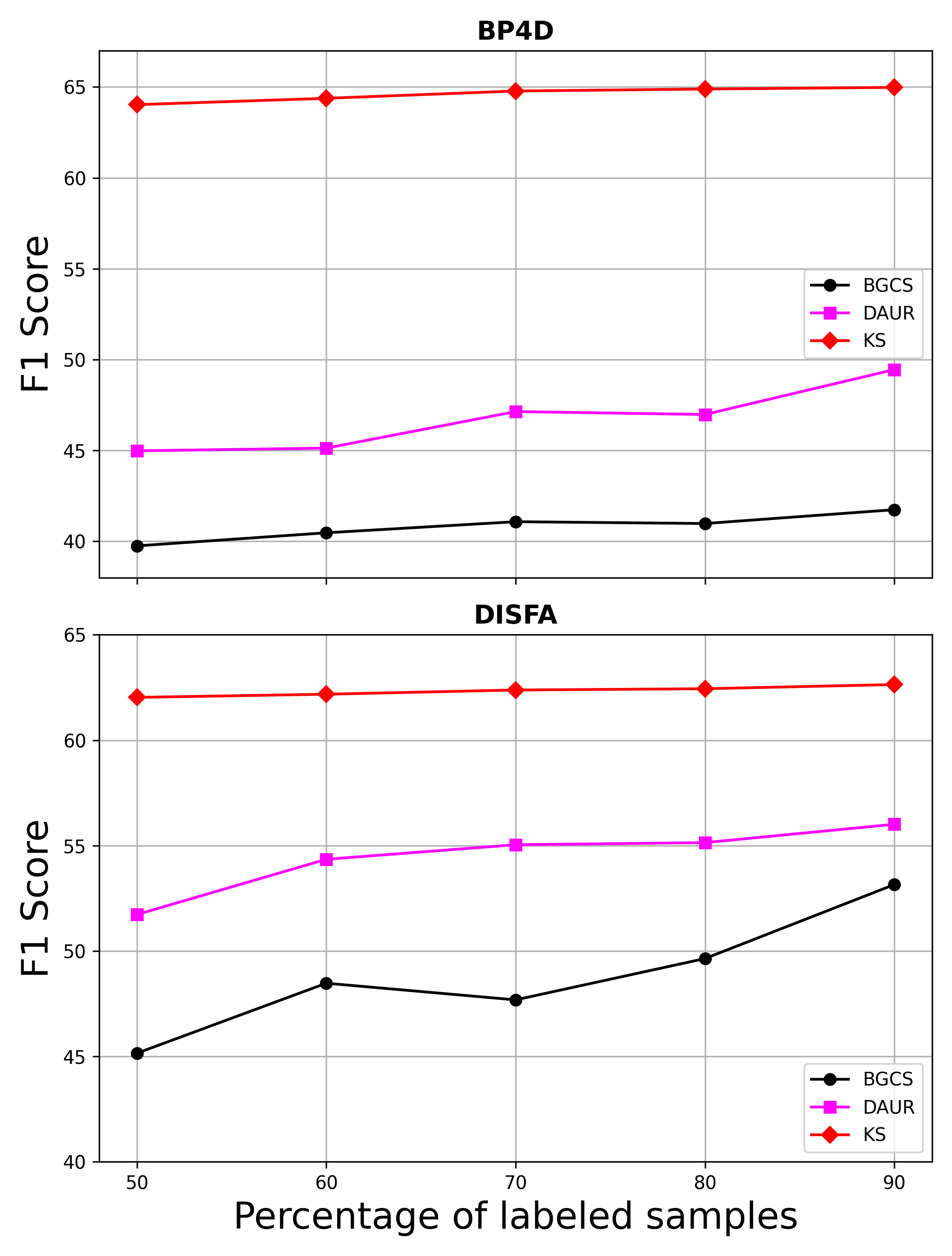}
\vspace{-2mm}
\caption{F1 Score of AU detection methods with incomplete annotations on BP4D and DISFA datasets.} 
\label{SSLonAU}
\end{figure}


To compare methods, the annotations of datasets are further modified to match the corresponding task. For the classification task, the performance of expressions and AUs are expressed in terms of accuracy and F1-score, respectively. For a fair comparison, we have chosen methods that follow the same experimental protocol for expressions and AUs and presented in Tables \ref{Results on Classification of expressions} and \ref{Results on Classification of action units}, respectively. Unless specified, all the results are shown in a within-database setting.

\noindent \textbf{Inexact Annotations:}
Although the UNBC-McMaster dataset is annotated with pain intensity levels, it has been used for classification by converting its ordinal labels (OPI ratings) to binary labels based on a threshold, i.e., OPI $\geq$ 3 is considered as pain and OPI$=$0 no pain, resulting in a total of 149 sequences with 57 positive and 92 negative bags. In all these methods, Leave-One-Subject-Out (LOSO) cross-validation is followed. For AU classification, BP4D and DISFA are widely used, where 8 AUs are considered for DISFA and 12 AUs for BP4D. For the DISFA dataset, AU intensities equal or greater than 2 are considered as occurrences, otherwise considered as non-occurrences. The DISFA dataset has a more significant data imbalance problem than BP4D, in which most of the AUs have very low occurrence rates.    


\noindent \textbf{Incomplete Annotations:}
RAF-DB and FER+ are widely used datasets for expression classification with SSL. SSL methods for FER have been evaluated using two experimental setups. First, the labeled images are randomly chosen at growing levels of supervision at 100, 400, 1000, 2000 and 4000 within RAF-DB and FER+, where the remaining images are considered to be unlabeled. Since most approaches \cite{9880204,9629313,Florea_eccv2020} use both Resnet-18 \cite{7780459} and WideResNet-28-2 \cite{DBLP:conf/bmvc/ZagoruykoK16} as backbones pre-trained on MS-Celeb-1M \cite{10.1007/978-3-319-46487-9_6}, we have compared methods on both backbones (see Figure \ref{fig:SSLonExp}). Second, some methods \cite{9629313,ijcai2025p154,10955730,9880204} explored training with the complete labeled images of RAF-DB and FER+, while AffectNet images are used as unlabeled images. For AU detection, BP4D and DISFA also widely rely on two experimental setups. Some of the methods \cite{7163081,Li_2023_ICCV,8237688} explored within-dataset evaluation, where the number of available labels is randomly dropped with a growing proportion of 50, 60, 70, 80 and 90 as shown in Figure \ref{SSLonAU}. In contrast, some of the methods \cite{NIPS2019_8377,9736614} used the EmotionNet dataset as unlabeled images, while the complete set of labels of within-data are used.

\noindent \textbf{{Inaccurate Annotations:}}
RAF-DB, FER+, and AffectNet have been widely used to deal with noisy annotations for expression recognition. For the AffectNet dataset, 33,803 facial images are considered clean data, and 1,71,005 facial images are considered noisy data for training. 3500 facial images are further used for validation, with 500 images per expression. In the case of RAF-DB, 12271 training samples are used for training, and the results are reported on the test set of 3068 samples. For the FER+ dataset, 28,709 images are used for training, 3,589 for validation, and 3,589 for testing. With all these datasets, the results are shown based on the training and testing partitions provided by the dataset organizers.

\noindent \textbf{Indirect Annotations:}
For expression classification, the MAFW dataset has been widely explored for indirect supervision, as it is the only dataset with textual descriptions for the facial expressions. For AU classification, 7319 frames are chosen from 30 video sequences of 17 subjects that exhibit the expression of pain with PSPI $\geq$ 5. Six AU labels are associated with the chosen frames, i.e., AU4, AU6, AU7, AU9, AU10, and AU43, which are related to pain expressions. In all these methods, 5-fold cross-validation is used. 
For CK+, MMI, and DISFA datasets, AUs, which are available for more than 10\% of all frames, are chosen. Based on this criterion,  
309 sequences of 106 subjects are chosen from 593 sequences of 123 subjects, resulting in 12 AUs for the CK+ dataset. 
For the MMI dataset, AUs are available for more than 10\% of all samples, resulting in 171 sequences from 27 subjects with 13 labels. For the DISFA dataset, 482 apex frames are chosen based on AU intensity levels, for which expression labels are obtained by FACS. Similar to CK+ and MMI, 9 AUs are considered. 5-fold cross-validation is deployed, where 20\% of the whole database is used as a validation set according to subjects. All the experiments are conducted in a subject-independent protocol.

\subsection{Critical Analysis and Results}

\subsubsection{Temporal Modeling}
For FER, temporal modeling has been explored primarily under inexact and indirect supervision. Under inexact supervision, early approaches aggregated frame evidence via max-pooling \cite{SIKKA2014659} or leveraged facial landmark displacements \cite{7163116}. Subsequent work of LOMo \cite{7780971} encoded temporal progression with ordinal constraints, yielding clear gains over conventional aggregates \cite{SIKKA2014659,7163116}. More recently, \cite{10204168} jointly captured short- and long-term dynamics using R3D features with multi-head attention, further improving robustness. Under indirect supervision, EmoCLIP \cite{foteinopoulou_emoclip_2024} enhanced CLIP image embeddings with transformer-based temporal modeling and reported a substantial performance margin (about 6\%).

For AU recognition under WSL, most prior work emphasizes modeling inter–AU relationships, while temporal cues remain comparatively underexplored. Notable exceptions under incomplete annotations include KS \cite{Li_2023_ICCV}, which models temporal evolution in the student with transformers and propagates temporal context to refine the teacher, exhibiting strong label-efficiency (minimal degradation as the labeled fraction decreases) and reporting an absolute improvement of roughly 15\% on BP4D (see Fig \ref{SSLonAU}). Similarly, WSRTL \cite{9736614} incorporates optical flow as an auxiliary objective to inject motion information, yielding consistent gains of about 5\% across both datasets of BP4D and DISFA (see Table \ref{Results on Classification of action units}). These results indicate that explicit temporal modeling complements relational AU reasoning and is particularly beneficial when annotations are sparse.


\subsubsection{Mining AU relationships}
Effective modeling of intricate inter-AU dependencies is pivotal for AU recognition under weak supervision, as AUs exhibit structured co-occurrence and mutual exclusivity patterns. Early approaches \cite{8329513,8237688} explored RBMs for modeling inter-AU relationships due to the inherent ability of capturing higher dependencies across multiple AUs, regularizing missing supervision. Recently, SMA-Net \cite{10154146} addressed attention collapse by encouraging each head to encode distinct AU-related features, promoting representational diversity within the same spatial regions. 
This internal diversification reduces redundancy among heads and improves the model’s ability to capture co-occurring or subtle AUs under weak supervision, thereby outperforming prior baselines under inexact supervision (about 4\%).
In the case of incomplete supervision, a subset of methods \cite{NIPS2019_8377,9736614,Li_2023_ICCV} explicitly models AU relationships, while others focus on learning more discriminative AU features \cite{10095708,WU20152279} (implicitly capturing dependencies). MLCR \cite{NIPS2019_8377} uses GCNs to encode dynamic AU co-activation graphs, propagating information among related AUs. WSRTL \cite{9736614} adopts a transformer encoder–decoder design in which the encoder refines AU-aware context and the decoder retrieves AU-specific cues via AU queries. Complementary to these, KS \cite{Li_2023_ICCV} exploits key frames to encode AU relationships using transformers along with temporal context to guide the student model, demonstrating that relational priors and temporal cues are synergistic when labels are sparse. Overall, transformer-based relational encoders provide a flexible and empirically strong mechanism for encoding inter–AU structure under WSL, offering significant improvement over prior baselines. Another promising direction is to jointly model expression–AU relations along with inter–AU dependencies for indirect supervision. 
\cite{7410779} explored expression-AU mappings with individual AUs, whereas subsequent works \cite{8329513,8578331,8578634} emphasized modeling both expression-dependent and expression-independent AUs and showed consistent performance improvement on multiple datasets. Recent methods employed dual/auxiliary-task objectives to implicitly strengthen consistency of expression-AU relationships \cite{8712447}, achieving state-of-the-art results under indirect supervision.

\subsubsection{Pseudo-label Refinement}
Ensuring pseudo-label reliability is key to achieving robust performance when annotations are sparse. CFRN \cite{10219749} exploits CAMs to detect spurious pseudo-labels under the premise that incorrect labels induce fake saliency, achieving state-of-the-art results on RAF-DB (see Fig.~\ref{fig:SSLonExp}). Beyond CAM-based refinement, recent methods \cite{ijcai2025p154,10955730} target pseudo-label noise from the perspectives of class imbalance and domain shift. ReMoP \cite{ijcai2025p154} explored consistency with a mixture of model predictions to identify unreliable pseudo-labels and mitigate confirmation bias. DDL \cite{10955730} learns adaptive, class-aware attention constraints calibrated by a balanced validation set to restore minority-class recall. Both approaches deliver comparable gains and achieve state-of-the-art performance on both RAF-DB and FER+, underscoring that reliability mechanisms, whether saliency, ensemble, or calibration-driven are crucial to standard semi-supervised pipelines.

\subsubsection{Uncertainty Estimation}
Discriminating clean from noisy labels and quantifying label uncertainty are central challenges for inaccurate supervision. Confidence-based ranking (SCN) \cite{9157210} prioritizes putatively clean samples, while relative (sample-wise) uncertainty estimation (RUL) \cite{zhang2021relative} improves the detection of ambiguous labels. Complementarily, class-specific uncertainty modeling \cite{10614222} accounts for varying uncertainty levels across classes. Beyond hard one-hot targets, assigning soft label distributions mitigates memorization of noisy labels. DMUE \cite{9578636} estimated reliable label distributions by aggregating diverse predictors using multi–auxiliary classifiers. Confidence-aware label smoothing (SLS) \cite{10.1007/978-3-031-19775-8_38} tempers targets using prediction confidence, yielding a modest improvement over \cite{9578636} on RAF-DB and comparable performance on AffectNet and FER+. Another line of research to deal with uncertain labels is ensemble-based approaches. Model ensembling (IPA2LT) \cite{10.1007/978-3-030-01261-8_14} aggregates predictions across multiple networks to mitigate the influence of unstable labels, while GAAVE \cite{Zheng_Li_Zhang_Wu_Cao_Ding_2023} explored adversarial-sensitivity analysis to flag uncertain samples by measuring prediction inconsistencies under small, targeted perturbations. Similar to the idea of exploiting CAMs \cite{10219749} to detect fake pseudo-labels, EAC \cite{10.1007/978-3-031-19809-0_24} enforced attention consistency across strong augmentations to deal with uncertain labels, achieving the best performance across AffectNet and FER+ datasets.

\section{Methods for Regression} \label{WSL for Regression}


Since regression in FABA under weak supervision has primarily been studied in the context of inexact and incomplete supervision, we organize this section around expression and AU intensity estimation under these two scenarios. 



\subsection{Inexact Annotations} \label{Inexact annotations for regression}



\subsubsection{Expression Intensity Estimation} \label{Inexact annotations for regression for Expressions}
Early works on weakly supervised expression intensity estimation framed the problem as mapping a global ordinal label (e.g., low–medium–high pain) to latent frame-level intensities. Ruiz et al. \cite{10.1007/978-3-319-54184-6_11} introduced Multi-instance Dynamic Ordinal Random Fields (MI-DORF), which treats frame-level pain intensity as an unobserved ordinal variable constrained by the video-level label. The model represents frame-level intensities as latent ordinal variables linked in a temporal sequence, and uses a structured probabilistic framework with high-order potentials to capture both temporal dynamics and ensure consistency with the global video-level label. This formulation enables instance-level inference without dense labels, but relies on hand-crafted features and strong distributional assumptions. Subsequently, 
Praveen et al. \cite{Praveen1} proposed a weakly supervised domain adaptation (WSDA) framework that combines MIL with adversarial domain adaptation. Using deep spatiotemporal features, the model transfers intensity knowledge from a fully annotated source dataset to a sparsely labeled target dataset. This approach shows that deep models benefit more from additional cross-dataset supervision than earlier feature-based methods, improving generalization across domains with different subjects and recording conditions.

\subsubsection{AU Intensity Estimation}
\label{Approaches for AU Intensity Estimation}
The MI-DORF framework \cite{8347018} was later extended to tackle AU intensity estimation, which introduced added challenge of modeling multiple AUs simultaneously, each with its own temporal dynamics and co-occurrence structure. The model treats frame-level AU intensities as latent ordinal variables and enforces consistency with the sequence-level label using maximum or relative aggregation functions. However, relying on global labels can lead to under-representation of more subtle or distributed intensity variations, especially when different AUs evolve asynchronously within a sequence. 
To address this limitation, Zhang et al. \cite{zhangbilateral} introduced Bilateral Ordinal Relevance Multiple Instance Regression (BORMIR), which injects limited domain knowledge by identifying both peak and valley frames. By jointly optimizing ordinal relevance, temporal smoothness, and relevance consistency, BORMIR captures the progressive nature of facial behavior more accurately than peak-only formulations. This highlights the importance of modeling intensity evolution, not just extremes, under weak supervision.

\subsubsection{Discussion}
Overall, ordinal MIL formulations provide a principled way to propagate global intensity annotations to instance-level predictions by exploiting ordering constraints \cite{10.1007/978-3-319-54184-6_11}. However, early approaches are constrained by hand-crafted representations and limited adaptability across datasets \cite{8347018,zhangbilateral}. Incorporating domain knowledge, such as peak–valley structure, improves discrimination between neighboring intensity levels \cite{zhangbilateral}, while leveraging deep representations and cross-dataset supervision substantially enhances robustness and generalization \cite{Praveen1}. These trends indicate a shift from static, feature-driven ordinal models toward data-driven, domain-adaptive frameworks that combine ordinal structure with learned representations to mitigate dataset bias and sparse annotation.

\subsection{Incomplete Annotations} \label{Incomplete for regression}


\subsubsection{Expression Intensity Estimation} \label{Incomplete Expression Intensity Labels for regression}
Directly addressing incomplete expression intensity annotations remains relatively underexplored. Rui et al. \cite{7780746} formulated the problem using ordinal support vector regression (OSVR) by explicitly modeling intensity levels as ordered categories rather than arbitrary numerical values. This design choice reflects the key intuition that relative ordering of intensities is more reliable than absolute intensity values under sparse supervision. By optimizing a max–margin objective that enforces ordinal consistency, OSVR enables stable learning even when annotations are irregular or missing. While the optimization is solved efficiently, the main contribution lies in preserving ordinal structure, which proves crucial for robust intensity estimation under weak supervision.



\subsubsection{AU Intensity Estimation} \label{Incomplete AU Intensity Estimation}
Most early works on AU intensity estimation under partial supervision rely on supervision based on keyframes, where a small number of peak and valley intensity frames serve as anchors for learning. The core assumption is that these frames provide reliable intensity cues, which can then be propagated to unlabeled frames by exploiting temporal continuity and facial structure. KBSS \cite{Zhang_2018_CVPR} exemplified this paradigm by transferring supervision from peak and valley frames using domain priors such as facial symmetry, monotonic intensity evolution over time, and appearance similarity between neighboring frames. These priors constrain the learning process, allowing the model to infer smooth intensity trajectories despite sparse annotations. Building on this idea, KJRE \cite{8954307} improved the performance by jointly learning both the facial representations and the intensity regression model, rather than treating the feature extractor as fixed. To compensate for the absence of dense labels, it incorporated various forms of weak supervision derived from human knowledge (feature smoothness, label consistency, etc) as soft or hard constraints. This hybrid supervision strategy reduces sensitivity to keyframe selection errors and improves generalization when annotations are irregularly distributed. Recognizing that different AUs exhibit distinct activation patterns and learning speeds, Zhang et al. \cite{9010955} proposed a context-aware fusion framework for weakly supervised AU intensity estimation that operates on facial patches instead of full-face images. A feature fusion module models spatial interactions by adapting each patch’s contribution based on the target AU, while a label fusion module captures AU-specific temporal dynamics. These attention-based modules enable the model to handle diverse activation patterns and support training with partially labeled data, improving robustness under limited supervision. 
More recently, Self-SL \cite{10.1007/978-3-030-69541-5_7} demonstrates that self-supervised pretraining on large unlabeled video corpora can provide strong spatiotemporal representations that already encode facial dynamics and temporal consistency. As a result, fine-tuning on sparsely labeled AU intensities requires fewer handcrafted priors or explicit keyframe annotations, shifting the burden from manual supervision design to data-driven representation learning.

\subsubsection{Discussion}
For expression intensity estimation, ordinal formulations such as OSVR \cite{7780746} highlight the importance of respecting label ordering when supervision is incomplete, offering stable learning across heterogeneous annotation densities. In AU intensity estimation, early reliance on peak and valley frames provides strong but sparse supervisory signals \cite{Zhang_2018_CVPR}, while domain priors—facial symmetry, temporal smoothness, and relative intensity ordering—play a central role in propagating reliable predictions to unlabeled frames \cite{Zhang_2018_CVPR,8954307}. Later work emphasizes AU-specific learning dynamics and inter-AU structure, improving calibration and plausibility under partial labels \cite{9010955,8705351}. Self-supervised video pretraining \cite{10.1007/978-3-030-69541-5_7} emerges as a scalable alternative that shifts reliance away from handcrafted priors toward data-driven representations. The field is moving from tightly constrained, keyframe-driven supervision toward representation-centric and pretraining-based approaches that better accommodate sparse, irregular, and incomplete intensity annotations.

\begin{table*}[!t]
\renewcommand{\arraystretch}{1.8}
\caption{Performance of state-of-art methods for regression of expressions and AUs under various modes of WSL setting on the most widely evaluated datasets.}
\label{Results on Regression}
\centering
\resizebox{\textwidth}{!}{
\begin{tabular}{|c|c|c|c|c|c|c|c|c|c|}
\hline
\textbf{WSL Setting} & \textbf{Dataset} &\textbf{Method} & \textbf{Task} & \textbf{Features} & \textbf{Learning Model} & \textbf{Validation} & \textbf{MAE} & \textbf{PCC} & \textbf{ICC} \\ 
\hline
\hline
\multirow{8}{*}{Inexact} & \multirow{4}{*}{UNBC-McMaster} & Ruiz et al. \cite{8347018} (TIP 2018) & Pain [6 levels]  & geometric & DORF & LOSO & 0.710 & 0.360 & 0.340\\ \cline{3-10}
                            &   & Praveen et al. \cite{Praveen1} (FG 2020) & Pain [6 levels] & I3D \cite{8099985} & WSDA & LOSO & 0.714 & 0.630 & 0.567\\ \cline{3-10}
                             &   & Praveen et al. \cite{RAJASEKHAR2021104167} (IMAVIS 2020) & Pain [6 levels] & I3D \cite{8099985} & WSDA-OR & LOSO & 0.530 & 0.705 & 0.696\\ \cline{3-10}
                             & & \multirow{1}{*}{ Zhang et al. \cite{zhangbilateral} (CVPR 2018)} & 
                            \multirow{1}{*}{Pain [6 levels]} & \multirow{1}{*}{geometric} & \multirow{1}{*}{BORMIR} & LOSO &  \multirow{1}{*}{0.821} & \multirow{1}{*}{0.605} & \multirow{1}{*}{0.531}\\ \cline{2-10}
& \multirow{3}{*}{DISFA} & Ruiz et al. \cite{8347018} (TIP 2018)  & 12 AUs [6 levels]  & geometric & DORF & 5-fold & 1.130 & 0.400 & 0.260\\ \cline{3-10}
                            &   & Zhao et al. \cite{7780746} (CVPR 2016) & 12 AUs [6 levels] & LBP & OSVR & 5-fold &  1.380 & 0.350 & 0.150\\ \cline{3-10} 
                            & & \multirow{1}{*}{Zhang et al. \cite{zhangbilateral} (CVPR 2018)} &
                            12 AUs [6 levels] & \multirow{1}{*}{geometric} & \multirow{1}{*}{BORMIR} & 5-fold &  \multirow{1}{*}{0.789} & \multirow{1}{*}{0.353} & \multirow{1}{*}{0.283}\\ \cline{2-10} 
                             
& \multirow{1}{*}{FERA 2015} &  Zhang et al. \cite{zhangbilateral} (CVPR 2018) & 5 AUs [6 levels] & geometric & BORMIR & Valster et al. \cite{7284874} &  0.852 & 0.635 & 0.620\\ \cline{2-10} \hline \hline  
\multirow{17}{*}{Incomplete} & \multirow{2}{*}{UNBC-McMaster} & Ruiz et al. \cite{8347018} (TIP 2018)  & Pain [6 levels]  & geometric & DORF & LOSO & 0.510 & 0.460 & 0.460 \\ \cline{3-10}
                            & & Zhao et al. \cite{7780746} (CVPR 2016) & Pain [6 levels] & LBP & OSVR & LOSO &  0.951 & 0.544 & 0.495 \\ \cline{2-10}
&\multirow{7}{*}{DISFA} & Ruiz et al. \cite{8347018} (TIP 2018)  & 12 AUs [6 levels]  & geometric & DORF & 5-fold & 0.480 & 0.420 & 0.380\\ \cline{3-10} 
 &   & Zhao et al. \cite{7780746} (CVPR 2016) & 12 AUs [6 levels] & LBP & OSVR & 5-fold &  0.800 & 0.370 & 0.290\\ \cline{3-10}
 &   & Zhang et al. \cite{Zhang_2018_CVPR} (CVPR 2018) & 12 AUs [6 levels] & CNN & KBSS & 3-fold &  0.330 & - & 0.360\\ \cline{3-10}
    &   & Wang et al. \cite{8705351} (TAFFC 2019) & 12 AUs[6 levels]  & RBM & RBM-P & 3-fold &  0.431 & 0.592 & 0.549\\  \cline{3-10}
    &   & Zhang et al. \cite{8954307} (CVPR 2019) & 12 AUs [6 levels] & geometric & KJRE & 5-fold &  0.910 & 0.370 & 0.350\\ 
                            \cline{3-10}                   
&   & Zhang et al. \cite{9010955} (ICCV 2019) & 12 AUs [6 levels] & Resnet-18 & CFLF & 3-fold &  0.329 & - & 0.408\\ \cline{3-10}
 &   & Sanchez et al. \cite{10.1007/978-3-030-69541-5_7} (ACCV 2020) & 12 AUs [6 levels] & Resnet-18 & Self-SL & 3-fold &  0.376 & - & 0.413\\ \cline{2-10}

& \multirow{6}{*}{FERA 2015} & Wang et al. \cite{8705351} (TAFFC 2019) & 5 AUs [6 levels]  & RBM & RBM-P & Valstar et al. \cite{7284874} &  0.728 & 0.605 & 0.585\\ \cline{3-10}
&   & Zhao et al. \cite{7780746} (CVPR 2016) & 5 AUs [6 levels] & LBP & OSVR & Valstar et al. \cite{7284874} &  1.077 & 0.545 & 0.544\\ \cline{3-10}
&   & Zhang et al. \cite{Zhang_2018_CVPR} (CVPR 2018) & 5 AUs [6 levels] & CNN & KBSS & Valstar et al. \cite{7284874} &  0.660 & - & 0.670\\ \cline{3-10}
 &   &  Zhang et al. \cite{8954307} (CVPR 2019) & 5 AUs [6 levels] & geometric & KJRE & Valstar et al. \cite{7284874} &  0.870 & 0.620 & 0.600\\ 
\cline{3-10}     
&   & Sanchez et al. \cite{10.1007/978-3-030-69541-5_7} (ACCV 2020) & 12 AUs [6 levels] & Resnet-18 & Self-SL & 3-fold &  0.798 & - & 0.680\\ \cline{3-10}
 &   & Zhang et al. \cite{9010955} (ICCV 2019) & 5 AUs [6 levels] & Resnet-18 & CFLF & Valstar et al. \cite{7284874} &  0.741 & - & 0.661\\ \hline 
\end{tabular}
}
\footnotetext{Footnote}
\end{table*}

\subsection{Model Evaluation Protocol} \label{exp for ordinal regression}



The intensities of pain and AUs are evaluated in terms of Mean Absolute Error (MAE), Pearson Correlation Coefficient (PCC), and Intra-class Correlation Coefficient (ICC). The comparison of performances of state-of-the-art methods, that follow the same experimental protocol for each category of WSL for regression is shown in Table \ref{Results on Regression}.    

\noindent \textbf{Inexact Annotations:}
In case of regression on the UNBC-McMaster dataset, PSPI labels of the frames are converted to 6 ordinal levels: 0(0), 1(1), 2(2), 3(3), 4-5(4), 6-15(5). The bag label of each pain sequence is considered the maximum of frame labels. Out of 25 subjects, 15 are used for training, 9 for validation, and 1 for testing. LOSO cross-validation is followed. For the DISFA and FERA datasets, bags are considered sequences with monotonically increasing or decreasing offset and onset frames. Similar to UNBC, the bag label is considered the maximum of the frame labels of AU intensities. 5-fold cross-validation is deployed for DISFA and official training/validation split for FERA. In order to compare the work of \cite{8347018} with the conventional approach of pain detection \cite{SIKKA2014659}, MILBOOST is deployed and the output probabilities of pain detection are normalized between 0 and 5 to have a fair comparison with that of \cite{8347018}.      

\noindent \textbf{Incomplete Annotations:} For the UNBC-McMaster dataset, only  10\% of annotations are considered in each sequence for the task of pain regression. 
For the task of AU regression in DISFA datasets, only 10\% of annotated frames are considered in \cite{8347018}, \cite{8705351}, and \cite{7780746} in order to incorporate the setting of incomplete annotations, whereas \cite{8867879,9010955,8954307} and \cite{Zhang_2018_CVPR} considered only the annotations of peak and valley frames, \cite{10.1007/978-3-030-69541-5_7} uses only 2\% of randomly annotated frames. In the case of the FERA 2015 dataset, official training and development sets provided by the FERA 2015 challenge \cite{7284874} are deployed. Similar to UNBC-McMaster and DISFA, \cite{8705351}, \cite{WANG201778} considers only 10\% of annotated frames while \cite{7780746}, \cite{zhangbilateral}, \cite{Zhang_2018_CVPR} considers annotations of peak and valley.

\subsection{Critical Analysis and Results}
Estimating the intensity of expressions or AUs is intrinsically harder than classification, because models must capture fine-grained, temporally evolving appearance changes and cope with label uncertainty. So, these regressions tasks are relatively under-explored. Under inexact supervision (sequence-level labels only), effective designs impose ordinal/MIL structure to map global labels to frame/segment intensities. Classical models encode temporal ordinal relations and smoothness constraints often via peak/valley anchors \cite{zhangbilateral,8347018}. Subsequent methods showed superior performance by employing deep spatiotemporal backbones that pair MIL with domain adaptation and ordinal priors \cite{Praveen1,RAJASEKHAR2021104167}.

For incomplete annotations, \cite{Zhang_2018_CVPR} and \cite{8954307} explored temporal, feature, and label smoothness in addition to the temporal relevance, where the former shows better performance by leveraging more constraints. \cite{9010955} explored attention mechanisms at both the feature and label levels, showing an improvement over prior approaches \cite{Zhang_2018_CVPR,8954307}. \cite{10.1007/978-3-030-69541-5_7} and \cite{8705351} further improved performance by leveraging large-scale pre-trained models based on self-supervised learning and RBMs to model AU relationships, where the latter has shown better performance. Compared to classical ML models \cite{zhangbilateral,8347018}, DL models \cite{8705351,10.1007/978-3-030-69541-5_7} have shown significant improvements for both inexact and incomplete annotations. However, a challenge in using DL models for estimating the intensity of facial expressions or AUs is the requirement of a large number of representative samples with intensity annotations, which is costly and requires domain expertise. Estimation of intensity levels using DL models is therefore explored in the context of incomplete annotations by leveraging large-scale pre-trained models \cite{10.1007/978-3-030-69541-5_7} or large-scale unlabeled data \cite{8867879}. 

\section{Challenges and Opportunities} 
\label{Challenges and Opportunities}

This section highlights key open challenges for FABA under weak supervision and outlines promising research opportunities to address them. While robust FER systems face many general hurdles (e.g., identity bias, data sparsity), we focus here on challenges unique to weakly annotated scenarios and how they may be addressed in future work.

\subsection{Few-Shot Learning} 
FSL offers a principled way to adapt models to new or rare expression categories from only a handful of labeled examples \cite{LU2023109480}. Recently, FSL has also gained attention in FER as a solution to the problem of novel classes for compound emotion recognition \cite{Zou2022WhenFE,CHEN2023206,zou2022learn}.
Xinyi et al.  \cite{Zou2022WhenFE} proposed a novel emotion-guided similarity network to deal with the limited annotations of compound expressions, where the pre-trained model is generalized to limited samples of novel classes (support set) of compound emotions. Zou et al. \cite{zou2022learn} also addressed the problem of compound FER in a cross-domain setting using a cascaded decomposition network for FSL. Chen et al. \cite{CHEN2023206} explored self-supervised vision transformers by jointly pertaining with multiple pretext tasks, and leveraged FSL to train the deep model with fewer labeled samples. Despite these advances in fully supervised settings, FSL remains an under-explored problem for weakly annotated FABA. For instance, incomplete supervision with severe class imbalance can be reframed as a few-shot problem by treating minority classes as new categories. By training on the majority classes and unlabeled data using semi-supervised strategies, a model could then adapt to the new minority classes, expanding the space for modeling mixed emotional states. This approach offers a novel way to mitigate confirmation bias and reduce spurious pseudo-labels. More broadly, coupling FSL with WSL provides a unified framework to address both limited/weak annotations and limited target-domain data.

\subsection{Multimodal Learning}

Exploring multiple modalities can reduce annotation cost and error by providing complementary cues \cite{10.3389/fdata.2020.00019}. To foster the progress, several multimodal datasets (audio, visual, text) have been released \cite{6553805,Kollias2019,7173007,6617456}, and many supervised works study audio–visual fusion for emotion recognition \cite{8070966,9667055,10005783,10095234}. Text has recently gained traction as a surrogate for human annotations \cite{Zhao_2025_WACV}, yet other modalities—such as physiology and speech—remain underexplored in the WSL setting. Thermal imagery offers invariance to illumination and can complement RGB for FER \cite{8253869,Pan:2018:FER:3240508.3240608}. Using 3D/4D facial data can help address pose and occlusion, capturing fine-grained AU changes (e.g., AU18 vs.\ AU10+AU17+AU24) via depth cues, see \cite{SANDBACH2012683}. Optical flow provides motion-sensitive descriptors of facial muscle activity by converting temporal changes into static textures \cite{7286757,6854261}, with efficacy shown for both macro- and micro-expressions \cite{Allaert2017ConsistentOF}. Therefore, modalities such as text, audio, optical flow, depth, and thermal images are promising auxiliary signals to supplement weak supervision. They can supply proxy labels, stabilize localization, and improve robustness under missing, inexact, or noisy annotations, laying a path toward resilient multimodal FABA under WSL.


\subsection{Fairness in Weakly Supervised Learning}

Fairness and weak annotation are intrinsically coupled in FABA. Recent studies demonstrated that many FER datasets contain significant annotation biases between genders, especially between happy and angry expressions \cite{10.1145/3531146.3533159,9710276}. Though the issue of fairness has been explored with supervised learning settings \cite{10582007,9710276}, it remains under-explored in the context of WSL. Weak labels (incomplete, inexact, or noisy) are rarely distributed uniformly across demographic groups. Source biases arising from annotator perspectives, cultural differences, varying lighting and camera domains, and demographic imbalances frequently result in group-dependent label noise. A model can inadvertently amplify this noise, especially when employing confidence-driven pseudo-labeling or MIL strategies that disproportionately favor already well-represented groups. This critical linkage motivates several promising directions, such as (i) uncertainty-aware, fairness-calibrated pseudo-labeling with group-normalized thresholds, (ii) design fairness-aware MIL/localization that balances attention across skin tones, ages, and genders, and (iii) incorporates causal/counterfactual augmentation (e.g., edit pose/illumination/skin tone while holding expression constant) to probe and reduce spurious correlations.  Therefore, this line of research holds significant promise for advancing both the robustness and ethical integrity of FABA under WSL.


\subsection{Micro-Expression Recognition Under Weak Supervision} 
Micro-expressions are brief, involuntary facial movements that reflect concealed or unconscious emotions, typically lasting less than 500 milliseconds \cite{9915437}. Their subtlety and sparsity make them particularly challenging to annotate at frame-level granularity, especially in unconstrained or real-world settings. Consequently, WSL presents a compelling alternative by reducing dependence on dense annotations and leveraging indirect or auxiliary signals. Recent studies have explored the close correlation between action units (AUs) and micro-expressions \cite{LI2021221}, suggesting that AU activations can serve as effective implicit supervision signals for micro-expression recognition. Robust temporal modeling is critical for capturing the fine-grained dynamics of micro-expressions in videos, particularly under weak supervision \cite{8373898}. Techniques such as spatiotemporal contrastive learning have demonstrated the ability to distinguish subtle muscle activations from background noise or overlapping macro-expressions \cite{10521757}. Furthermore, cross-domain transfer from macro-expression datasets or AU activation patterns may enhance generalization, especially when guided by domain adaptation or multi-task objectives. The integration of auxiliary tasks such as macro-expression recognition offers additional supervision pathways to compensate for weak visual signals \cite{ijcai2021p164}. Emerging efforts also leverage multimodal large language models (MLLMs) to fuse human-like perception of subtle facial movements with powerful semantic reasoning capabilities \cite{zhang2025mellmexploringllmpoweredmicroexpression}. To advance the field, there is a need for new benchmark datasets incorporating coarse, probabilistic, or confidence-weighted annotations, as well as evaluation protocols tailored to the weak supervision regime. Progress in this direction holds the potential to enable robust and fine-grained micro-expression recognition systems, particularly for real-time applications.

\subsection{Leveraging Large Language Models} 
LLMs have strong potential to enhance FABA by compensating for weak or noisy supervision, injecting linguistic structure, expert priors, and explanatory rationales into the training pipeline. Acting as label refiners, LLMs transform low-quality annotations, such as clip-level tags, coarse timestamps, or ASR transcripts, into candidate expression or AU labels, complete with textual rationales and uncertainty scores that can serve as soft targets or consistency constraints for vision models. This process can be informed by encoded expert knowledge, like display rules and AU–expression taxonomies, to probabilistically fuse multiple weak signals. Recently, LLM-derived annotations have been successfully explored for indirect supervision for FER, circumventing the need for human annotations \cite{Zhao_2025_WACV}. In multimodal settings, LLM-derived summaries provide temporal priors that guide attention mechanisms toward relevant frames, while also supporting data-centric curation efforts by generating descriptive prompts for diffusion models to synthesize diverse, label-coherent faces, and rewriting crowd-sourced tags into calibrated, subgroup-aware pseudo-labels. Furthermore, LLMs enable explanation-aware supervision by verifying whether model saliency aligns with natural language descriptions of AUs, triggering reweighting or active learning queries when rationales misalign, thus regularizing MIL and enabling text–video contrastive training. Future directions and evaluation must consider not only accuracy but also rationale faithfulness, localization, calibration, and subgroup performance gaps, while mitigating cultural bias and hallucination via self-consistency checks, verifier prompts, and robust human-in-the-loop audits.

\subsection{Towards Unified and Interpretable Models}
Most current methods tackle weakly supervised classification, regression, or AU detection in isolation. We argue for a unified framework that natively accommodates heterogeneous supervision—categorical, ordinal, and continuous—under incomplete, inexact, or noisy labels. Such a system would eliminate isolated pipelines, improve robustness, and generalize better across datasets and tasks. Concretely, the framework should enable multi-task learning (expressions, AUs, valence–arousal), cross-modal reasoning (video, audio, text/ASR), and interpretable outputs (region/temporal rationales and concept bottlenecks). Building on strong pretrained backbones, it can use modular adapters such as LoRA for rapid task/domain personalization, and integrate uncertainty estimation to gate pseudo-labels, calibrate decisions, and handle label noise. Training this architecture at scale by pooling datasets spanning multiple weak-supervision regimes (MIL, sparse/clip-level tags, noisy multi-annotator labels) would yield a foundation model for affective behavior, later adapted with lightweight heads/adapters to diverse downstream settings—reducing annotation cost while delivering consistent, explainable performance across tasks, subjects, and domains.

\section{Conclusion}
This survey provides a structured review of WSL methods for FABA. To better understand the field, we categorize the reviewed literature based on affective tasks and WSL scenarios. We systematically analyze state-of-the-art methods for recognizing expressions and AUs under various WSL conditions, discussing their challenges and strengths. By comparing results from widely used experimental protocols, we provide new insights into the benefits and limitations of these approaches. This review highlights the diverse WSL techniques available for developing robust deep learning models for real-world FABA applications. We also present open challenges and suggest future research directions to advance the field. This survey serves as a comprehensive reference for researchers and practitioners, fostering progress in FABA under WSL settings.

\ifCLASSOPTIONcaptionsoff
  \newpage
\fi



\bibliographystyle{IEEEtran}
\bibliography{references}

@ARTICLE{10154146,
  author={Li, Xiaotian and Zhang, Zheng and Zhang, Xiang and Wang, Taoyue and Li, Zhihua and Yang, Huiyuan and Ciftci, Umur and Ji, Qiang and Cohn, Jeffrey and Yin, Lijun},
  journal={IEEE Trans. on Affective Computing, \textnormal{15:2, 620-631}}, 
  title={Disagreement Matters: Exploring Internal Diversification for Redundant Attention in Generic Facial Action Analysis}, 
  year={2024},
}

@inproceedings{10.5555/3295222.3295349,
author = {Vaswani, Ashish and Shazeer, Noam and Parmar, Niki and Uszkoreit, Jakob and Jones, Llion and Gomez, Aidan N. and Kaiser, \L{}ukasz and Polosukhin, Illia},
title = {Attention is all you need},
year = {2017},
booktitle = {NeurIPS},
}

@InProceedings{pmlr-v139-radford21a,
  title = 	 {Learning Transferable Visual Models From Natural Language Supervision},
  author =       {Radford, Alec and Kim, Jong Wook and Hallacy, Chris and Ramesh, Aditya and Goh, Gabriel and Agarwal, Sandhini and Sastry, Girish and Askell, Amanda and Mishkin, Pamela and Clark, Jack and Krueger, Gretchen and Sutskever, Ilya},
  booktitle = 	 {ICML},
  year = 	 {2021},
}

@ARTICLE{7439855,
  author={Geng, Xin},
  journal={IEEE Tran. on Knowledge and Data Engineering \textnormal{28:7, 1734-1748}}, 
  title={Label Distribution Learning}, 
  year={2016},
}

@inproceedings{10.1145/1273496.1273596,
author = {Salakhutdinov, Ruslan and Mnih, Andriy and Hinton, Geoffrey},
title = {Restricted Boltzmann machines for collaborative filtering},
year = {2007},
booktitle = {ICML},
}

@inproceedings{
laine2017temporal,
title={Temporal Ensembling for Semi-Supervised Learning},
author={Samuli Laine and Timo Aila},
booktitle={ICLR},
year={2017},
}

@inproceedings{Lee2013PseudoLabelT,
  title={Pseudo-label: The simple and efficient semi-supervised learning method for deep neural networks},
  author={Lee, Dong-Hyun and others},
  booktitle={Workshop on challenges in representation learning, ICML},
  year={2013},
}

@ARTICLE{8880512,
  author={Shao, Zhiwen and Liu, Zhilei and Cai, Jianfei and Wu, Yunsheng and Ma, Lizhuang},
  journal={IEEE Trans. on Affective Computing, \textnormal{13:3, 1274-1289}}, 
  title={Facial Action Unit Detection Using Attention and Relation Learning}, 
  year={2022},
  }

@INPROCEEDINGS{10204168,
  author={Wang, Hanyang and Li, Bo and Wu, Shuang and Shen, Siyuan and Liu, Feng and Ding, Shouhong and Zhou, Aimin},
  booktitle={CVPR}, 
  title={Rethinking the Learning Paradigm for Dynamic Facial Expression Recognition}, 
  year={2023},
  volume={},
  number={},
}

@article{zhang2025mellmexploringllmpoweredmicroexpression,
  title={{MELLM: Exploring LLM-Powered Micro-Expression Understanding Enhanced by Subtle Motion Perception}},
  author={Zhang, Zhengye and Zhao, Sirui and Liu, Shifeng and Yin, Shukang and Mao, Xinglong and Xu, Tong and Chen, Enhong},
  journal={arXiv:2505.07007},
  year={2025}
}

@INPROCEEDINGS{8373898,
  author={Khor, Huai-Qian and See, John and Phan, Raphael Chung Wei and Lin, Weiyao},
  booktitle={FG}, 
  title={Enriched Long-Term Recurrent Convolutional Network for Facial Micro-Expression Recognition}, 
  year={2018},
  volume={},
  number={},
}

@article{LI2021221,
title = {Micro-expression action unit detection with spatial and channel attention},
journal = {Neurocomputing, \textnormal{436, 221-231}},
year = {2021},
author = {Yante Li and Xiaohua Huang and Guoying Zhao},
}

@inproceedings{NIPS1997_82965d4e,
 author = {Maron, Oded and Lozano-P\'{e}rez, Tom\'{a}s},
 booktitle = {NeurIPS},
 pages = {},
 title = {A Framework for Multiple-Instance Learning},
 year = {1997}
}

@ARTICLE{9915437,
  author={Li, Yante and Wei, Jinsheng and Liu, Yang and Kauttonen, Janne and Zhao, Guoying},
  journal={IEEE Trans. on Affective Computing, \textnormal{13:4, 2028-2046}}, 
  title={Deep Learning for Micro-Expression Recognition: A Survey}, 
  year={2022},
}

@article{FE,
author = {Shichuan Du  and Yong Tao  and Aleix M. Martinez },
title = {Compound facial expressions of emotion},
journal = {Proc. of the National Academy of Sciences, \textnormal{111, E1454-E1462}},
year = {2014},
}

@ARTICLE{8974606,
  author={Wang, Kai and Peng, Xiaojiang and Yang, Jianfei and Meng, Debin and Qiao, Yu},
  journal={IEEE Trans. on Image Processing, \textnormal{29, 4057-4069}}, 
  title={Region Attention Networks for Pose and Occlusion Robust Facial Expression Recognition}, 
  year={2020},
}

@INPROCEEDINGS{9956496,
  author={Bonnard, Jules and Dapogny, Arnaud and Dhombres, Ferdinand and Bailly, Kevin},
  booktitle={ICPR}, 
  title={Privileged Attribution Constrained Deep Networks for Facial Expression Recognition}, 
  year={2022},
  volume={},
  number={},
}

@ARTICLE{10521757,
  author={Bao, Yongtang and Wu, Chenxi and Zhang, Peng and Shan, Caifeng and Qi, Yue and Ben, Xianye},
  journal={IEEE Trans. on Affective Computing, \textnormal{15:4, 2083--2096}}, 
  title={Boosting Micro-Expression Recognition via Self-Expression Reconstruction and Memory Contrastive Learning}, 
  year={2024},
}

@inproceedings{ijcai2021p164,
  title     = {Micro-Expression Recognition Enhanced by Macro-Expression from Spatial-Temporal Domain},
  author    = {Xia, Bin and Wang, Shangfei},
  booktitle = {IJCAI},
  year      = {2021},
}

@ARTICLE{de2020deep,
  author={de Melo, Wheidima Carneiro and Granger, Eric and Hadid, Abdenour},
  journal={IEEE Trans. on Affective Computing \textnormal{13:3, 1581--1592}}, 
  title={A Deep Multiscale Spatiotemporal Network for Assessing Depression From Facial Dynamics}, 
  year={2022},
  }

@INPROCEEDINGS{8954307,
  author={Zhang, Yong and Wu, Baoyuan and Dong, Weiming and Li, Zhifeng and Liu, Wei and Hu, Bao-Gang and Ji, Qiang},
  booktitle={CVPR}, 
  title={Joint Representation and Estimator Learning for Facial Action Unit Intensity Estimation}, 
  year={2019},
  volume={},
  number={},
 }

@InProceedings{zou2022learn,
author="Zou, Xinyi
and Yan, Yan
and Xue, Jing Hao
and Chen, Si
and Wang, Hanzi",
title="Learn-to-Decompose: Cascaded Decomposition Network for Cross-Domain Few-Shot Facial Expression Recognition",
booktitle="ECCV",
year="2022",
}

@article{CHEN2023206,
title = {Self-supervised vision transformer-based few-shot learning for facial expression recognition},
journal = {Information Sciences, \textnormal{634, 206-226}},
year = {2023},
author = {Xuanchi Chen and Xiangwei Zheng and Kai Sun and Weilong Liu and Yuang Zhang},
}

@book{Mehrabian,
  author    = {Albert Mehrabian},
  title     = {Nonverbal Communication},
  publisher = {Routledge},
  year      = {2017}
}

@article{ZHANG2014692,
title = "{BP4D}-{Spontaneous}: a high-resolution spontaneous {3D} dynamic facial expression database",
journal = "Image and Vision Computing, \textnormal{32:10, 692 - 706}",
year = "2014",
issn = "0262-8856",
doi = "https://doi.org/10.1016/j.Image and Vision Computing.2014.06.002",
author = "Xing Zhang and Lijun Yin and Jeffrey F. Cohn and Shaun Canavan and Michael Reale and Andy Horowitz and Peng Liu and Jeffrey M. Girard",
}

@Article{Kollias2019,
author="Kollias, Dimitrios
and Tzirakis, Panagiotis
and Nicolaou, Mihalis A.
and Papaioannou, Athanasios
and Zhao, Guoying
and Schuller, Bj{\"o}rn
and Kotsia, Irene
and Zafeiriou, Stefanos",
title="{Deep Affect Prediction in-the-Wild: Aff-Wild Database and Challenge, Deep Architectures, and Beyond}",
journal="IJCV, \textnormal{127, 907-929}",
year="2019",
}

@INPROCEEDINGS{10219749,
  author={Sun, Hao and Pi, Chenchen and Xie, Wei},
  booktitle={ICME}, 
  title={Semi-Supervised Facial Expression Recognition by Exploring False Pseudo-Labels}, 
  year={2023},
  volume={},
  number={},
}

@ARTICLE{8013713, 
author={A. {Mollahosseini} and B. {Hasani} and M. H. {Mahoor}}, 
journal={IEEE Trans. on Affective Computing, \textnormal{10:1, 18-31}}, 
title={{AffectNet: A Database for Facial Expression, Valence, and Arousal  Comp. in the Wild}}, 
year={2019}, 
}

@INPROCEEDINGS{5543262,
  author={Lucey, Patrick and Cohn, Jeffrey F. and Kanade, Takeo and Saragih, Jason and Ambadar, Zara and Matthews, Iain},
  booktitle={CVPRW}, 
  title={{The Extended Cohn-Kanade Dataset (CK+): A complete dataset for action unit and emotion-specified expression}}, 
  year={2010},
  volume={},
  number={},
}

@INPROCEEDINGS{1521424,
  author={Pantic, M. and Valstar, M. and Rademaker, R. and Maat, L.},
  booktitle={IEEE ICME}, 
  title={Web-based database for facial expression analysis}, 
  year={2005},
  volume={},
  number={},
}

@InProceedings{FER2013,
author="Goodfellow, Ian J.
and Erhan, Dumitru
and Carrier, Pierre Luc
and Courville, Aaron
and Mirza, Mehdi
and Hamner, Ben
and Cukierski, Will
and Tang, Yichuan
and Thaler, David
and Lee, Dong-Hyun
and Zhou, Yingbo
and Ramaiah, Chetan
and Feng, Fangxiang
and Li, Ruifan
and Wang, Xiaojie
and Athanasakis, Dimitris
and Shawe-Taylor, John
and Milakov, Maxim
and Park, John
and Ionescu, Radu
and Popescu, Marius
and Grozea, Cristian
and Bergstra, James
and Xie, Jingjing
and Romaszko, Lukasz
and Xu, Bing
and Chuang, Zhang
and Bengio, Yoshua",
title="Challenges in Representation Learning: A Report on Three Machine Learning Contests",
booktitle="NeurIPS",
year="2013",
}

@INPROCEEDINGS{8237690,
  author={Benitez-Quiroz, C. Fabian and Wang, Yan and Martinez, Aleix M.},
  booktitle={ICCV}, 
  title={Recognition of Action Units in the Wild with Deep Nets and a New Global-Local Loss}, 
  year={2017},
  volume={},
  number={},
}

@INPROCEEDINGS{8099985,
  author={J. {Carreira} and A. {Zisserman}},
  booktitle={CVPR}, 
  title={Quo Vadis, Action Recognition? A New Model and the Kinetics Dataset}, 
  year={2017},
  volume={},
  number={},

}

@INPROCEEDINGS{7780969, 
author={C. F. {Benitez-Quiroz} and R. {Srinivasan} and A. M. {Martinez}}, 
booktitle={CVPR}, 
title={EmotioNet: An Accurate, Real-Time Algorithm for the Automatic Annotation of a Million Facial Expressions in the Wild}, 
year={2016}, 
volume={}, 
number={}, 
}

@INPROCEEDINGS{Zhao_2018_CVPR,
  author={Zhao, Kaili and Chu, Wen-Sheng and Martinez, Aleix M.},
  booktitle={CVPR}, 
  title={Learning Facial Action Units from Web Images with Scalable Weakly Supervised Clustering}, 
  year={2018},
  volume={},
  number={},
}

@INPROCEEDINGS{7780971, 
author={K. Sikka and G. Sharma and M. Bartlett}, 
booktitle={CVPR}, 
title={{LOMo: Latent Ordinal Model for Facial Analysis in Videos}}, 
year={2016}, 
volume={}, 
number={}, 

}

@article{SAADI2024122784,
title = {Driver’s facial expression recognition: A comprehensive survey},
journal = {Expert Systems with Applications \textnormal{242, 122784}},
year = {2024},
author = {Ibtissam Saadi and Douglas W. Cunningham and Abdelmalik Taleb-Ahmed and Abdenour Hadid and Yassin El Hillali},
}

@inproceedings{Thomas:2018:PEI:3242969.3264984,
 author = {Thomas, Chinchu and Nair, Nitin and Jayagopi, Dinesh Babu},
 title = {Predicting Engagement Intensity in the Wild Using Temporal Convolutional Network},
 booktitle = {ACM ICMI},
 year = {2018},
}

@INPROCEEDINGS{Praveen1,
  author={R. Gnana Praveen and Granger, Eric and Cardinal, Patrick},
  booktitle={FG}, 
  title={Deep Weakly Supervised Domain Adaptation for Pain Localization in Videos}, 
  year={2020},
  volume={},
  number={},
}

@INPROCEEDINGS{10095234,
  author={R. Gnana Praveen and Granger, Eric and Cardinal, Patrick},
  booktitle={IEEE ICASSP}, 
  title={Recursive Joint Attention for Audio-Visual Fusion in Regression Based Emotion Recognition}, 
  year={2023},
  volume={},
  number={},
}

@ARTICLE{10005783,
  author={R. Gnana Praveen and Cardinal, Patrick and Granger, Eric},
  journal={IEEE Trans. on Biometrics, Behavior, and Identity Science, \textnormal{5:3, 360-373}}, 
  title={Audio–Visual Fusion for Emotion Recognition in the Valence–Arousal Space Using Joint Cross-Attention}, 
  year={2023},
}

@INPROCEEDINGS{8237688,
  author={Wu, Shan and Wang, Shangfei and Pan, Bowen and Ji, Qiang},
  booktitle={ICCV}, 
  title={Deep Facial Action Unit Recognition from Partially Labeled Data}, 
  year={2017},
  volume={},
  number={},
}

@ARTICLE{8705351,
  author={Wang, Shangfei and Pan, Bowen and Wu, Shan and Ji, Qiang},
  journal={IEEE Trans. on Affective Computing, \textnormal{12:4, 1018-1030}}, 
  title={Deep Facial Action Unit Recognition and Intensity Estimation from Partially Labelled Data}, 
  year={2021},
 }

@INPROCEEDINGS{7284874,
  author={Valstar, Michel F. and Almaev, Timur and Girard, Jeffrey M. and McKeown, Gary and Mehu, Marc and Yin, Lijun and Pantic, Maja and Cohn, Jeffrey F.},
  booktitle={FG}, 
  title={{FERA 2015} - second Facial Expression Recognition and Analysis challenge}, 
  year={2015},
}

@ARTICLE{5959155, 
author={G. {McKeown} and M. {Valstar} and R. {Cowie} and M. {Pantic} and M. {Schroder}}, 
journal={IEEE Trans. on Affective Computing, \textnormal{3:1, 5-17}}, 
title={{The SEMAINE Database: Annotated Multimodal Records of Emotionally Colored Conversations between a Person and a Limited Agent}}, 
year={2012}, 
}

@INPROCEEDINGS{7780746, 
author={R. {Zhao} and Q. {Gan} and S. {Wang} and Q. {Ji}},
booktitle={CVPR}, 
title={Facial Expression Intensity Estimation Using Ordinal Information}, 
year={2016}, 
volume={}, 
number={}, 

}

@article{WANG201778,
title = {Expression-assisted facial action unit recognition under incomplete AU annotation},
journal = {Pattern Recognition, \textnormal{61, 78-91}},
year = {2017},
author = {Shangfei Wang and Quan Gan and Qiang Ji},
}

@INPROCEEDINGS{7163081,
  author={Song, Yale and McDuff, Daniel and Vasisht, Deepak and Kapoor, Ashish},
  booktitle={FG}, 
  title={Exploiting sparsity and co-occurrence structure for action unit recognition}, 
  year={2015},
}

@ARTICLE{7308029, 
author={M. X. Huang and G. Ngai and K. A. Hua and S. C. F. Chan and H. V. Leong}, 
journal={IEEE Trans. on Affective Computing, \textnormal{7:4, 360-373}}, 
title={Identifying User-Specific Facial Affects from Spontaneous Expressions with Minimal Annotation}, 
year={2016}, 
}

@INPROCEEDINGS{9157210,
  author={Wang, Kai and Peng, Xiaojiang and Yang, Jianfei and Lu, Shijian and Qiao, Yu},
  booktitle={CVPR}, 
  title={Suppressing Uncertainties for Large-Scale Facial Expression Recognition}, 
  year={2020},
  volume={},
  number={},
}

@inproceedings{
zhang2021relative,
title={Relative Uncertainty Learning for Facial Expression Recognition},
author={Yuhang Zhang and Chengrui Wang and Weihong Deng},
booktitle={NeurIPS},
year={2021},
}

@INPROCEEDINGS{9578636,
  author={She, Jiahui and Hu, Yibo and Shi, Hailin and Wang, Jun and Shen, Qiu and Mei, Tao},
  booktitle={CVPR}, 
  title={Dive into Ambiguity: Latent Distribution Mining and Pairwise Uncertainty Estimation for Facial Expression Recognition}, 
  year={2021},
  volume={},
  number={},
}

@article{Zhao_Liu_Zhou_2021, 
title={Robust Lightweight Facial Expression Recognition Network with Label Distribution Training}, 
journal={AAAI}, 
author={Zhao, Zengqun and Liu, Qingshan and Zhou, Feng}, year={2021}, 
}

@inproceedings{10.1145/3503161.3547960,
author = {Shao, Jianjian and Wu, Zhenqian and Luo, Yuanyan and Huang, Shudong and Pu, Xiaorong and Ren, Yazhou},
title = {Self-Paced Label Distribution Learning for In-The-Wild Facial Expression Recognition},
year = {2022},
booktitle = {ACM MM},
}

@INPROCEEDINGS{zhangbilateral,
  author={Zhang, Yong and Zhao, Rui and Dong, Weiming and Hu, Bao-Gang and Ji, Qiang},
  booktitle={CVPR}, 
  title={Bilateral Ordinal Relevance Multi-instance Regression for Facial Action Unit Intensity Estimation}, 
  year={2018},
  volume={},
  number={},
}

@ARTICLE{8347018, 
author={A. {Ruiz} and O. {Rudovic} and X. {Binefa} and M. {Pantic}}, 
journal={IEEE Trans. on Image Processing, \textnormal{27:8, 3969-3982}}, 
title={Multi-Instance Dynamic Ordinal Random Fields for Weakly Supervised Facial Behavior Analysis}, 
year={2018}, 
}

@article{WSL,
author = {Zhou, Zhi-Hua},
title = {A brief introduction to weakly supervised learning},
journal = {National Science Review, \textnormal{5:1, 44-53}},
year = {2018},
}

@article{ZHANG2025131409,
title = {A survey on self-supervised learning: Recent advances and open problems},
journal = {Neurocomputing, \textnormal{655, 131409}},
year = {2025},
author = {Jianping Zhang and Lei Yang and Seyed Mahmoud {Sajjadi Mohammadabadi} and Feng Yan},
}

@inproceedings{Barsoum:2016:TDN:2993148.2993165,
author = {Barsoum, Emad and Zhang, Cha and Ferrer, Cristian Canton and Zhang, Zhengyou},
title = {Training deep networks for facial expression recognition with crowd-sourced label distribution},
year = {2016},
booktitle = {ACM ICMI},
}

@article{peng2019dual, 
title={Dual Semi-Supervised Learning for Facial Action Unit Recognition}, 
journal={AAAI}, 
author={Peng, Guozhu and Wang, Shangfei}, 
year={2019}, 
}

@InProceedings{Florea_eccv2020,
author="Florea, Corneliu
and Badea, Mihai
and Florea, Laura
and Racoviteanu, Andrei
and Vertan, Constantin",
title="{Margin-Mix: Semi-Supervised Learning for Face Expression Recognition}",
booktitle="ECCV",
year="2020",
}

@InProceedings{10.1007/978-3-030-69541-5_7,
author="Sanchez, Enrique
and Bulat, Adrian
and Zaganidis, Anestis
and Tzimiropoulos, Georgios",
title="Semi-supervised Facial Action Unit Intensity Estimation with Contrastive Learning",
booktitle="ACCV",
year="2021",
}

@incollection{NIPS2019_8377,
title = {{Multi-label Co-regularization for Semi-supervised Facial  AU Recognition}},
author = {Niu, Xuesong and Han, Hu and Shan, Shiguang and Chen, Xilin},
booktitle = {NeurIPS},
year = {2019},
}

@INPROCEEDINGS{8099760,
  author={Li, Shan and Deng, Weihong and Du, JunPing},
  booktitle={CVPR}, 
  title={Reliable Crowdsourcing and Deep Locality-Preserving Learning for Expression Recognition in the Wild}, 
  year={2017},
  volume={},
  number={},
}

@article{CARBONNEAU,
author = "Marc Andre Carbonneau and Veronika Cheplygina and Eric Granger and Ghyslain Gagnon",
title = "Multiple instance learning: A survey of problem characteristics and applications",
journal = "Pattern Recognition, \textnormal{77, 329 - 353}",
year = "2018",
}

@INPROCEEDINGS{9667055,
  author={Praveen, R. Gnana and Granger, Eric and Cardinal, Patrick},
  booktitle={FG}, 
  title={Cross Attentional Audio-Visual Fusion for Dimensional Emotion Recognition}, 
  year={2021},
  volume={},
  number={},
}

@article{RAJASEKHAR2021104167,
title = {Deep domain adaptation with ordinal regression for pain assessment using weakly-labeled videos},
journal = {Image and Vision  Comp., \textnormal{110, 104167}},
year = {2021},
author = {R. Gnana Praveen and Eric Granger and Patrick Cardinal},
}

@book{Ekman1978,
    author = {Ekman, Paul and Friesen, Wallace},
    title = {{Facial Action Coding System: A Technique for the Measurement of Facial Movement}},
    year = {1978},
    publisher = {Consulting Psychologists Press},
}

@article{POSNER_RUSSELL_PETERSON_2005, title={The circumplex model of affect: An integrative approach to affective neuroscience, cognitive development, and psychopathology}, journal={Development and Psychopathology, \textnormal{17:3, 715–734}}, 
author={Posner, JONATHAN and Russell, JAMES A. and Peterson, BRADLEY S.}, year={2005}, 
}

@InProceedings{10.1007/978-3-030-01261-8_14,
author="Zeng, Jiabei
and Shan, Shiguang
and Chen, Xilin",
title="Facial Expression Recognition with Inconsistently Annotated Datasets",
booktitle="ECCV",
year="2018",
}

@ARTICLE{8712447,
  author={Wang, Shangfei and Peng, Guozhu},
  journal={IEEE Trans. on Multimedia, \textnormal{21:12, 3218-3230}}, 
  title={Weakly Supervised Dual Learning for Facial Action Unit Recognition}, 
  year={2019},
}

@INPROCEEDINGS{8578634,
  author={Zhang, Yong and Dong, Weiming and Hu, Bao-Gang and Ji, Qiang},
  booktitle={CVPR}, 
  title={Classifier Learning with Prior Probabilities for Facial Action Unit Recognition}, 
  year={2018},
  volume={},
  number={},
}

@InProceedings{Li_2023_ICCV,
    author    = {Li, Xiaotian and Zhang, Xiang and Wang, Taoyue and Yin, Lijun},
    title     = {Knowledge-Spreader: Learning Semi-Supervised Facial Action Dynamics by Consistifying Knowledge Granularity},
    booktitle = {ICCV},
    year      = {2023},
}

@INPROCEEDINGS{10095708,
  author={Liu, Zhongling and Liu, Rujie and Shi, Ziqiang and Liu, Liu and Mi, Xiaoyu and Murase, Kentaro},
  booktitle={IEEE ICASSP}, 
  title={Semi-Supervised Contrastive Learning with Soft Mask Attention for Facial Action Unit Detection}, 
  year={2023},
  volume={},
  number={},
}

@ARTICLE{8329513,
  author={Wang, Shangfei and Peng, Guozhu and Ji, Qiang},
  journal={IEEE Trans. on Affective Computing, \textnormal{11:4, 640-652}}, 
  title={Exploring Domain Knowledge for Facial Expression-Assisted Action Unit Activation Recognition}, 
  year={2020},
}

@InProceedings{10.1007/978-3-319-46487-9_6,
author="Guo, Yandong
and Zhang, Lei
and Hu, Yuxiao
and He, Xiaodong
and Gao, Jianfeng",
title="{MS-Celeb-1M: A Dataset and Benchmark for Large-Scale Face Recognition}",
booktitle="ECCV",
year="2016",

}

@inproceedings{DBLP:conf/bmvc/ZagoruykoK16,
  author       = {Sergey Zagoruyko and
                  Nikos Komodakis},
  title        = {Wide Residual Networks},
  booktitle    = {BMVC},
  year         = {2016},
}

@INPROCEEDINGS{7780459, 
author={K. He and X. Zhang and S. Ren and J. Sun}, 
booktitle={CVPR}, 
title={Deep Residual Learning for Image Recognition}, 
year={2016}, 
volume={}, 
number={}, 
}

@article{Martinez2016,
author="Martinez, Brais
and Valstar, Michel F.",
title="Advances, Challenges, and Opportunities in Automatic Facial Expression Recognition",
year="2016",
journal="Advances in Face Detection and Facial Image Analysis, \textnormal{63-100}",
}

@article{FASEL2003259,
title = "Automatic facial expression analysis: a survey",
journal = "Pattern Recognition, \textnormal{36:1, 259 - 275}",
year = "2003",
author = "B. Fasel and Juergen Luettin",
}

@incollection{multiple-instance-boosting-for-object-detection,
title = {Multiple Instance Boosting for Object Detection},
author = {Paul A. Viola and John C. Platt and Cha Zhang},
booktitle = {NeurIPS},
year = {2006},
}

@book{Chapelle2010,
author = {Chapelle, Olivier and Schlkopf, Bernhard and Zien, Alexander},
title = {Semi-Supervised Learning},
year = {2010},
isbn = {0262514125},
publisher = {The MIT Press},
edition = {1st},
}

@INPROCEEDINGS{10582007,
  author={Sukumar, Aadith and Desai, Aditya and Singhal, Peeyush and Gokhale, Sai and Jain, Deepak Kumar and Walambe, Rahee and Kotecha, Ketan},
  booktitle={FG}, 
  title={Training Against Disguises: Addressing and Mitigating Bias in Facial Emotion Recognition with Synthetic Data}, 
  year={2024},
  volume={},
  number={},
}

@INPROCEEDINGS{9710276,
  author={Chen, Yunliang and Joo, Jungseock},
  booktitle={ICCV}, 
  title={Understanding and Mitigating Annotation Bias in Facial Expression Recognition}, 
  year={2021},
  volume={},
  number={},
}

@inproceedings{10.1145/3531146.3533159,
author = {Pahl, Jaspar and Rieger, Ines and M\"{o}ller, Anna and Wittenberg, Thomas and Schmid, Ute},
title = {Female, white, 27? Bias Evaluation on Data and Algorithms for Affect Recognition in Faces},
year = {2022},
booktitle = {ACM FAccT},
}

@inproceedings{Zou2022WhenFE,
  title={When Facial Expression Recognition Meets Few-Shot Learning: A Joint and Alternate Learning Framework},
  author={Xinyi Zou and Y. Yan and Jing-Hao Xue and Si Chen and Hanzi Wang},
  booktitle={AAAI},
  year={2022}
}

@ARTICLE{6940284, 
author={E. {Sariyanidi} and H. {Gunes} and A. {Cavallaro}}, 
journal={IEEE Trans. on Pattern Analysis and Machine Intelligence, \textnormal{37:6, 1113-1133}}, 
title={Automatic Analysis of Facial Affect: A Survey of Registration, Representation, and Recognition}, 
year={2015}, 
}

@ARTICLE{8928510,
  author={Hassan, Teena and Seuß, Dominik and Wollenberg, Johannes and Weitz, Katharina and Kunz, Miriam and Lautenbacher, Stefan and Garbas, Jens-Uwe and Schmid, Ute},
  journal={IEEE Trans. on Pattern Analysis and Machine Intelligence, \textnormal{43:6, 1815-1831}}, 
  title={Automatic Detection of Pain from Facial Expressions: A Survey}, 
  year={2021},
}

@article{CANAL2022593,
title = {A survey on facial emotion recognition techniques: A state-of-the-art literature review},
journal = {Information Sciences, \textnormal{582, 593-617}},
year = {2022},
author = {Felipe Zago Canal and Tobias Rossi Müller and Jhennifer Cristine Matias and Gustavo Gino Scotton and Antonio Reis {de Sa Junior} and Eliane Pozzebon and Antonio Carlos Sobieranski},
}

@ARTICLE{li2020deep,
  author={Li, Shan and Deng, Weihong},
  journal={IEEE Trans. on Affective Computing, \textnormal{13:3, 1195-1215}}, 
  title={Deep Facial Expression Recognition: A Survey}, 
  year={2022},
}

@Article{Matsumoto1992,
author="Matsumoto, David",
title="More evidence for the universality of a contempt expression",
journal="Motivation and Emotion, \textnormal{16, 363-368}",
year="1992",
}

@ARTICLE{10.3389/fdata.2020.00019,
AUTHOR={Lopez, Kevin and Fodeh, Samah J. and Allam, Ahmed and Brandt, Cynthia A. and Krauthammer, Michael},  
TITLE={Reducing Annotation Burden Through Multimodal Learning},  
JOURNAL={Frontiers in Big Data, \textnormal{3}},      
YEAR={2020}, 
}

@book{ekman76,
    author = {Ekman, Paul and Friesen, Wallace V.},
    location = {Palo Alto, CA},
    publisher = {Consulting Psychologists Press},
    title = {{Pictures of Facial Affect}},
    year = {1976}
}

@InProceedings{Wen,
author="Wen, Yandong
and Zhang, Kaipeng
and Li, Zhifeng
and Qiao, Yu",
title="A Discriminative Feature Learning Approach for Deep Face Recognition",
booktitle="ECCV",
year="2016",
}

@INPROCEEDINGS{6553805, 
author={F. {Ringeval} and A. {Sonderegger} and J. {Sauer} and D. {Lalanne}}, 
booktitle={FG}, 
title={{Introducing the RECOLA multimodal corpus of remote collaborative and affective interactions}}, 
year={2013}, 
volume={}, 
number={}, 
}

@ARTICLE{8867879,
  author={Zhang, Yong and Fan, Yanbo and Dong, Weiming and Hu, Bao-Gang and Ji, Qiang},
  journal={IEEE Access, \textnormal{7, 150743-150756}}, 
  title={Semi-Supervised Deep Neural Network for Joint Intensity Estimation of Multiple Facial Action Units}, 
  year={2019},
 }

@INPROCEEDINGS{9010955,
  author={Zhang, Yong and Jiang, Haiyong and Wu, Baoyuan and Fan, Yanbo and Ji, Qiang},
  booktitle={ICCV}, 
  title={Context-Aware Feature and Label Fusion for Facial Action Unit Intensity Estimation With Partially Labeled Data}, 
  year={2019},
  volume={},
  number={},
}

@InProceedings{10.1007/978-3-319-54184-6_11,
author="Ruiz, Adria
and Rudovic, Ognjen
and Binefa, Xavier
and Pantic, Maja",
title="Multi-Instance Dynamic Ordinal Random Fields for Weakly-Supervised Pain Intensity Estimation",
booktitle="ACCV",
year="2016",
}

@INPROCEEDINGS{SIKKA2014659,
  author={Sikka, Karan and Dhall, Abhinav and Bartlett, Marian},
  booktitle={FG}, 
  title={Weakly supervised pain localization using multiple instance learning}, 
  year={2013},
  volume={},
  number={},
}

@inproceedings{ijcai2020p145,
  title     = {Weakly Supervised Local-Global Relation Network for Facial Expression Recognition},
  author    = {Zhang, Haifeng and Su, Wen and Yu, Jun and Wang, Zengfu},
  booktitle = {IJCAI},
  year      = {2020},
}

@inproceedings{10.1145/3503161.3548190,
author = {Liu, Yuanyuan and Dai, Wei and Feng, Chuanxu and Wang, Wenbin and Yin, Guanghao and Zeng, Jiabei and Shan, Shiguang},
title = {{MAFW: A Large-scale, Multi-modal, Compound Affective Database for Dynamic Facial Expression Recognition in the Wild}},
year = {2022},
booktitle = {ACM MM},
}

@INPROCEEDINGS{10582016,
  author={Belharbi, Soufiane and Pedersoli, Marco and Koerich, Alessandro Lameiras and Bacon, Simon and Granger, Eric},
  booktitle={FG}, 
  title={Guided Interpretable Facial Expression Recognition via Spatial Action Unit Cues}, 
  year={2024},
  volume={},
  number={},
}

@INPROCEEDINGS{Zhang_2018_CVPR,
  author={Zhang, Yong and Dong, Weiming and Hu, Bao-Gang and Ji, Qiang},
  booktitle={CVPR}, 
  title={Weakly-Supervised Deep Convolutional Neural Network Learning for Facial Action Unit Intensity Estimation}, 
  year={2018},
  volume={},
  number={},
}

@inproceedings{10.1007/978-3-031-19775-8_38,
author = {Lukov, Tohar and Zhao, Na and Lee, Gim Hee and Lim, Ser-Nam},
title = {Teaching With Soft Label Smoothing For Mitigating Noisy Labels In Facial Expressions},
year = {2022},
booktitle = {ECCV}
}

@InProceedings{10.1007/978-3-031-19809-0_24,
author="Zhang, Yuhang
and Wang, Chengrui
and Ling, Xu
and Deng, Weihong",
title="Learn from All: Erasing Attention Consistency for Noisy Label Facial Expression Recognition",
booktitle="ECCV",
year="2022",
}

@INPROCEEDINGS{6617456,
  author={Walter, Steffen and Gruss, Sascha and Ehleiter, Hagen and Junwen Tan and Traue, Harald C. and Werner, Philipp and Al-Hamadi, Ayoub and Crawcour, Stephen and Andrade, Adriano O. and Moreira da Silva, Gustavo},
  booktitle={IEEE Int. Conf. on Cybernetics}, 
  title={The biovid heat pain database data for the advancement and systematic validation of an automated pain recognition system}, 
  year={2013},
  volume={},
  number={},
  }

@inproceedings{10.1145/3549865.3549908,
author = {Carreto Pic\'{o}n, Gin\'{e}s and Roig-Maim\'{o}, Maria Francesca and Mascar\'{o} Oliver, Miquel and Amengual Alcover, Esperan\c{c}a and Mas-Sans\'{o}, Ramon},
title = {Do Machines Better Understand Synthetic Facial Expressions than People?},
year = {2022},
booktitle = {Proc. of the XXII International Conference on Human Computer Interaction},
articleno = {18},
numpages = {5},
}

@InProceedings{Zhao_2025_WACV,
    author    = {Zhao, Zengqun and Cao, Yu and Gong, Shaogang and Patras, Ioannis},
    title     = {Enhancing Zero-Shot Facial Expression Recognition by LLM Knowledge Transfer},
    booktitle = {WACV},
    year      = {2025},
}

@inproceedings{48557,title	= {{MixMatch: A Holistic Approach to Semi-Supervised Learning}},author	= {David Berthelot and Nicholas Carlini and Ian Goodfellow and Nicolas Papernot and Avital Oliver and Colin Raffel},year	= {2019},booktitle	= {NeurIPS}}

@ARTICLE{9736614,
  author={Yan, Jingwei and Wang, Jingjing and Li, Qiang and Wang, Chunmao and Pu, Shiliang},
  journal={IEEE Trans. on Multimedia, \textnormal{25, 1760-1772}}, 
  title={Weakly Supervised Regional and Temporal Learning for Facial Action Unit Recognition}, 
  year={2023},
}

@inproceedings{foteinopoulou_emoclip_2024,
	title = {{EmoCLIP}: {A} {Vision}-{Language} {Method} for {Zero}-{Shot} {Video} {Facial} {Expression} {Recognition}},
	author = {Foteinopoulou, Niki Maria and Patras, Ioannis},
	year = {2024},
	booktitle={FG}
}

@inproceedings{zhang2023learning,
  title={Learning Emotion Representations from Verbal and Nonverbal Communication},
  author={Zhang, Sitao and Pan, Yimu and Wang, James Z},
  booktitle={CVPR},
  year={2023}
}

@inproceedings{Li_2023_BMVC,
author    = {Yifan Li and Hu Han and Shiguang Shan and zhilong ji and Jinfeng Bai and Xilin Chen},
title     = {{ReCoT:  Regularized Co-Training for Facial Action Unit Recognition with Noisy Labels}},
booktitle = {BMVC},
year      = {2023},
}

@ARTICLE{10614222,
  author={Nie, Wei and Wang, Zhiyong and Wang, Xinming and Chen, Bowen and Zhang, Hanlin and Liu, Honghai},
  journal={IEEE Trans. on Biometrics, Behavior, and Identity Science, \textnormal{7:1, 95-107}}, 
  title={Diving Into Sample Selection for Facial Expression Recognition With Noisy Annotations}, 
  year={2025},
}

@article{Zheng_Li_Zhang_Wu_Cao_Ding_2023, 
title={Attack Can Benefit: An Adversarial Approach to Recognizing Facial Expressions under Noisy Annotations}, 
journal={AAAI}, 
author={Zheng, Jiawen and Li, Bo and Zhang, Shengchuan and Wu, Shuang and Cao, Liujuan and Ding, Shouhong}, 
year={2023},  
}

@ARTICLE{1165342,
  author={Rabiner, L. and Juang, B.},
  journal={IEEE Signal Processing Magazine, \textnormal{3:1, 4-16}}, 
  title={{An introduction to hidden Markov models}}, 
  year={1986},
}

@inproceedings{MIR,
         author = "S. Ray and D. Page",
         title = "Multiple Instance Regression",
         booktitle =  "ICML",
         year = "2001"
         }

@article{LU2023109480,
title = {A survey on machine learning from few samples},
journal = {Pattern Recognition, \textnormal{139, 109480}},
year = {2023},
author = {Jiang Lu and Pinghua Gong and Jieping Ye and Jianwei Zhang and Changshui Zhang},
}

@ARTICLE{7173007, 
author={M. S. H. Aung and S. Kaltwang and B. Romera-Paredes and B. Martinez and A. Singh and M. Cella and M. Valstar and H. Meng and A. Kemp and M. Shafizadeh and A. C. Elkins and N. Kanakam and A. de Rothschild and N. Tyler and P. J. Watson and A. C. d. C. Williams and M. Pantic and N. Bianchi-Berthouze}, 
journal={IEEE Trans. on Affective Computing, \textnormal{7:4, 435-451}}, 
title={{The Automatic Detection of Chronic Pain-Related Expression: Requirements, Challenges and the Multimodal EmoPain Dataset}}, 
year={2016}, 
}

@inproceedings{Tran,
 author = {Tran, Du and Bourdev, Lubomir and Fergus, Rob and Torresani, Lorenzo and Paluri, Manohar},
 title = {{Learning Spatiotemporal Features with 3D Convolutional Networks}},
 booktitle = {ICCV},
 year = {2015},
}

@ARTICLE{8070966, 
author={P. Tzirakis and G. Trigeorgis and M. A. Nicolaou and B. W. Schuller and S. Zafeiriou}, 
journal={IEEE Journal of Selected Topics in Signal Processing, \textnormal{11:8, 1301-1309}}, 
title={End-to-End Multimodal Emotion Recognition Using Deep Neural Networks}, 
year={2017}, 
}

@inproceedings{Pan:2018:FER:3240508.3240608,
 author = {Pan, Bowen and Wang, Shangfei},
 title = {Facial Expression Recognition Enhanced by Thermal Images Through Adversarial Learning},
 booktitle = {ACM MM},
 year = {2018},
}

@inproceedings{Allaert2017ConsistentOF,
  title={Consistent Optical Flow Maps for Full and Micro Facial Expression Recognition},
  author={Benjamin Allaert and Ioan Marius Bilasco and Chabane Djeraba},
  booktitle={VISAPP},
  year={2017}
}

@INPROCEEDINGS{6854261, 
author={C. {Lee} and R. {Chellappa}}, 
booktitle={IEEE ICASSP}, 
title={Sparse localized facial motion dictionary learning for facial expression recognition}, 
year={2014}, 
volume={}, 
number={}
}

@ARTICLE{7286757, 
author={Y. {Liu} and J. {Zhang} and W. {Yan} and S. {Wang} and G. {Zhao} and X. {Fu}}, 
journal={IEEE Trans. on Affective Computing, \textnormal{7:4, 299-310}}, 
title={A Main Directional Mean Optical Flow Feature for Spontaneous Micro-Expression Recognition}, 
year={2016}, 
}

@article{SANDBACH2012683,
title = "Static and dynamic 3D facial expression recognition: A comprehensive survey",
journal = "Image and Vision  Comp. \textnormal{30:10, 683 - 697}",
year = "2012",
author = "Georgia Sandbach and Stefanos Zafeiriou and Maja Pantic and Lijun Yin",
}

@ARTICLE{8253869, 
author={S. {Wang} and B. {Pan} and H. {Chen} and Q. {Ji}}, 
journal={IEEE Tran. on Cybernetics, \textnormal{48:7, 2203-2214}}, 
title={Thermal Augmented Expression Recognition}, 
year={2018}, 
}

@INPROCEEDINGS{8578331, 
author={G. {Peng} and S. {Wang}}, 
booktitle={CVPR}, 
title={Weakly Supervised Facial Action Unit Recognition Through Adversarial Training}, 
year={2018}, 
volume={}, 
number={}, 
}

@ARTICLE{8472814, 
author={Wang, Shangfei and Peng, Guozhu and Chen, Shiyu and Ji, Qiang}, 
journal={IEEE Tran. on Cybernetics}, 
title={Weakly Supervised Facial Action Unit Recognition With Domain Knowledge
}, 
year={2018}, 
volume={48}, 
number={11}, 
pages={3265-3276}, 
}

@INPROCEEDINGS{7410779,
  author={Ruiz, Adria and Van De Weijer, Joost and Binefa, Xavier},
  booktitle={ICCV}, 
  title={From Emotions to Action Units with Hidden and Semi-Hidden-Task Learning}, 
  year={2015},
  volume={},
  number={},
}

@ARTICLE{10955730,
  author={Li, Yaqi and Jiang, Jing and Zhang, Yuhang and Fang, Han and Hu, Jiani and Deng, Weihong},
  journal={IEEE Trans. on Affective Computing, \textnormal{16:3, 2274-2286}}, 
  title={{DDL: Dynamic Direction Learning for Semi-Supervised Facial Expression Recognition}}, 
  year={2025},
  volume={},
  number={},
}

@inproceedings{ijcai2025p154,
  title     = {Towards Regularized Mixture of Predictions for Class-Imbalanced Semi-Supervised Facial Expression Recognition},
  author    = {Li, Hangyu and Zhang, Yixin and Yao, Jiangchao and Wang, Nannan and Han, Bo},
  booktitle = {IJCAI},
  year      = {2025},
}

@INPROCEEDINGS{9880204,
  author={Li, Hangyu and Wang, Nannan and Yang, Xi and Wang, Xiaoyu and Gao, Xinbo},
  booktitle={CVPR}, 
  title={Towards Semi-Supervised Deep Facial Expression Recognition with An Adaptive Confidence Margin}, 
  year={2022},
  volume={},
  number={},
}

@INPROCEEDINGS{10484312,
  author={Cho, Yunseong and Kim, Chanwoo and Cho, Hoseong and Ku, Yunhoe and Kim, Eunseo and Boboev, Muhammadjon and Lee, Joonseok and Baek, Seungryul},
  booktitle={WACV}, 
  title={{RMFER: Semi-supervised Contrastive Learning for Facial Expression Recognition with Reaction Mashup Video}}, 
  year={2024},
  volume={},
  number={},
}

@ARTICLE{9629313,
  author={Jiang, Jing and Deng, Weihong},
  journal={IEEE Trans. on Affective Computing, \textnormal{14:3, 2402-2414}}, 
  title={Boosting Facial Expression Recognition by A Semi-Supervised Progressive Teacher}, 
  year={2021},
}

@article{WU20152279,
title = {Multi-label learning with missing labels for image annotation and facial action unit recognition},
journal = {Pattern Recognition, \textnormal{48:7, 2279-2289}},
year = {2015},
author = {Baoyuan Wu and Siwei Lyu and Bao-Gang Hu and Qiang Ji},
}

@inproceedings{Adria,
	title = {{Regularized Multi-Concept MIL for weakly-supervised facial behavior categorization}},
	author = {Ruiz, Adria and Van de Weijer, Joost and Binefa, Xavier},
	year = {2014},
	booktitle = {BMVC},
	editors = {Valstar, Michel and French, Andrew and Pridmore, Tony}
}

@INPROCEEDINGS{7163116, 
author={C. Wu and S. Wang and Q. Ji}, 
booktitle={FG}, 
title={{Multi-instance Hidden Markov Model for facial expression recognition}}, 
year={2015}, 
number={}, 
}

@ARTICLE{6475933,
  author={Mavadati, S. Mohammad and Mahoor, Mohammad H. and Bartlett, Kevin and Trinh, Philip and Cohn, Jeffrey F.},
  journal={IEEE Trans. on Affective Computing, \textnormal{4:2, 151-160}}, 
  title={{DISFA}: A Spontaneous Facial Action Intensity Database}, 
  year={2013},
}

@inproceedings{5771462,
  author={Lucey, Patrick and Cohn, Jeffrey F. and Prkachin, Kenneth M. and Solomon, Patricia E. and Matthews, Iain},
  booktitle={FG}, 
  title={{Painful data: The UNBC-McMaster shoulder pain expression archive database}}, 
  year={2011},
  volume={},
  number={},
}
\vspace{-10mm}
\begin{IEEEbiography}[{\includegraphics[width=1in,height=1.25in,clip,keepaspectratio]{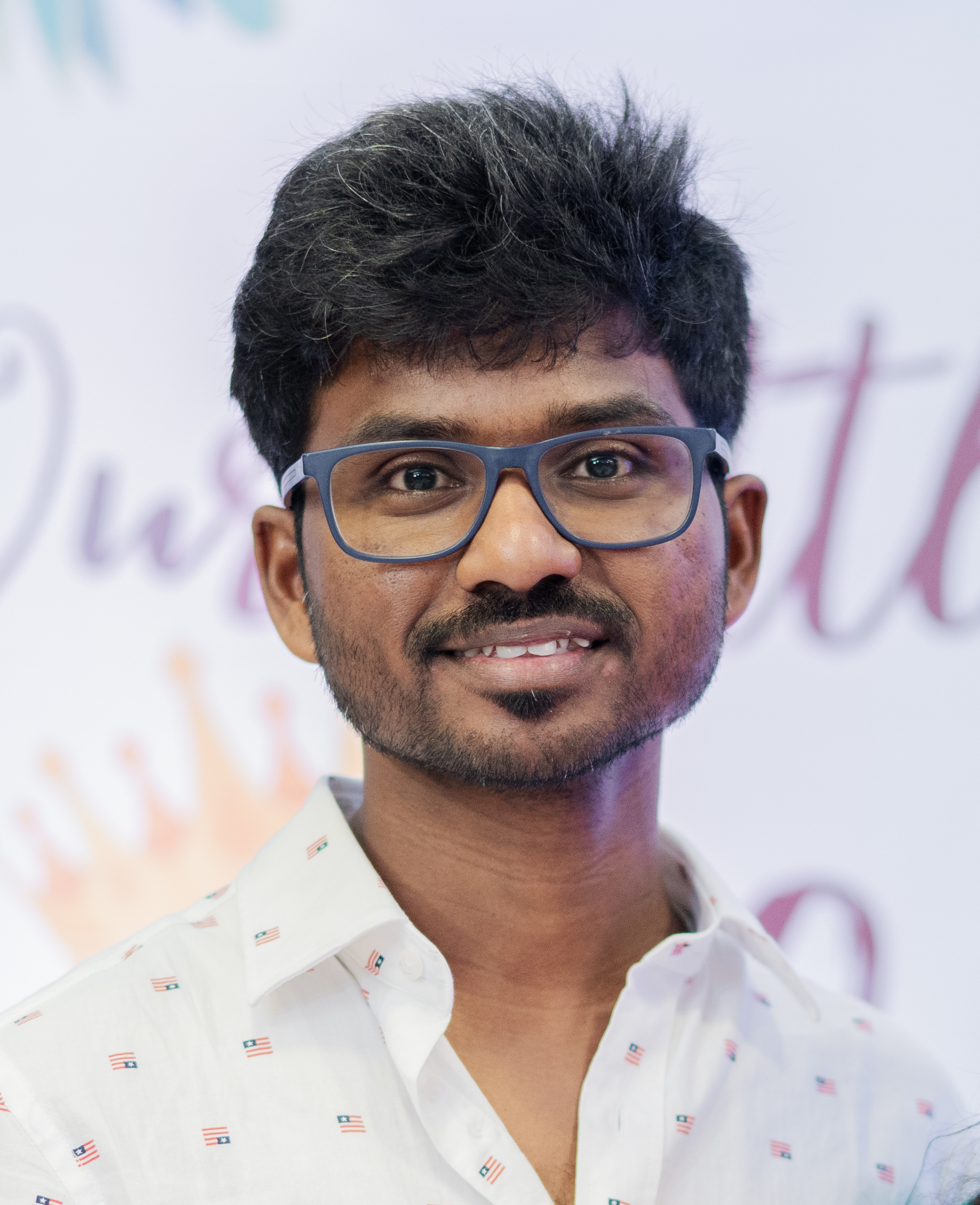}}]{R. Gnana Praveen}
received Master's degree from 
IITG, India in 2012, and Ph.D. degree from ETS Montreal, Canada in 2023. 
He also worked as a post-doctoral researcher at 
CRIM, Montreal on audio-visual learning for emotion recognition and speaker verification. He currently works as an AI researcher at Huawei Noah's Ark Lab in Montreal, Canada. 
His research interests include affective computing, machine/deep learning, computer vision, and multimodal learning. 
\end{IEEEbiography}
\vspace{-10mm}
\begin{IEEEbiography}[{\includegraphics[width=1in,height=1.25in,clip,keepaspectratio]{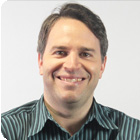}}]{Patrick Cardinal}
received 
M.Sc. degree from McGill University in 2003, and Ph.D. degree from ETS Montreal in 2013. From 2000 to 2013, he was involved in several projects related to speech processing, especially in the development of a closed-captioning system for live television shows based on automatic speech recognition. After his post-doctoral position at MIT, he joined ETS as a Professor. His research interests cover several aspects of speech processing for real-life and medical applications.
\end{IEEEbiography}
\vspace{-10mm}

\begin{IEEEbiography}[{\includegraphics[width=1in,height=1.25in,clip,keepaspectratio]{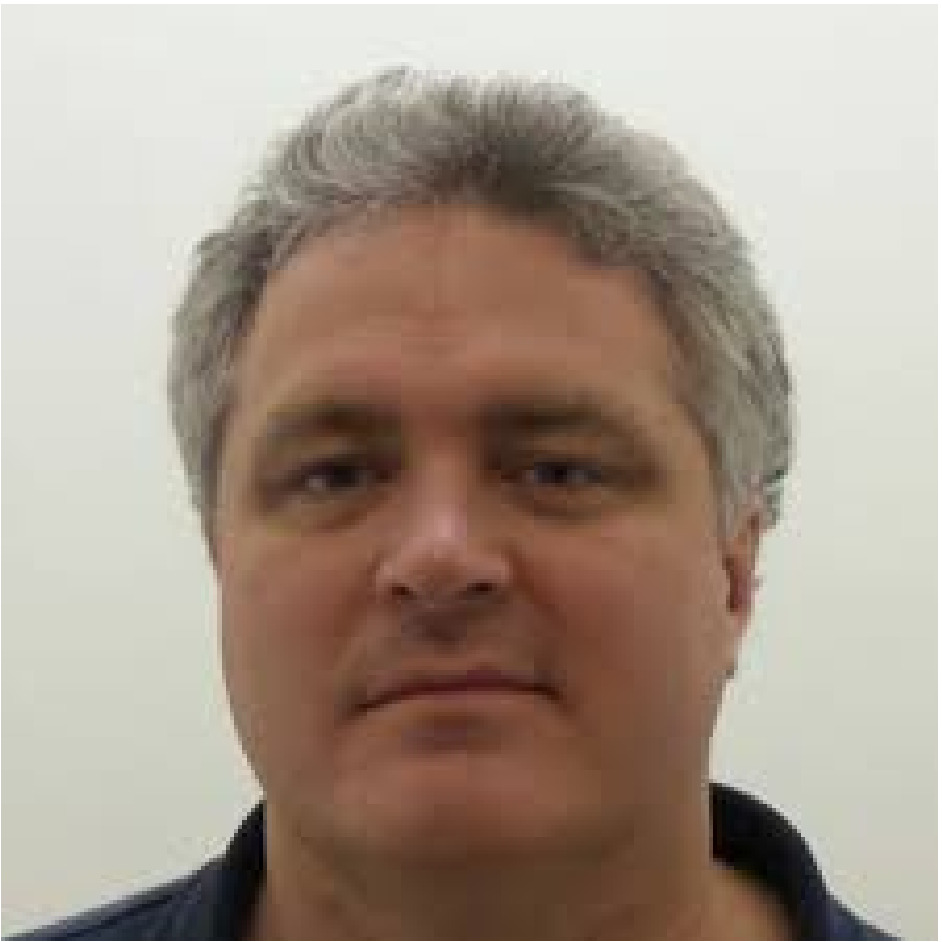}}]
{Eric Granger} received a Ph.D. degree in Electrical Engineering from Polytechnique Montréal in 2001. He was a Defense Scientist with DRDC Ottawa from 1999 to 2001, and in Research and Development with Mitel Networks from 2001 to 2004. He joined the Dept of Systems Engineering at ETS Montreal, Canada, in 2004, where he is currently a Full Professor and the Director of LIVIA, a research laboratory focused on computer vision and artificial intelligence. 
His research interests include pattern recognition, machine learning, information fusion, and computer vision, with applications in affective computing, biometrics, medical image analysis, and video analytics.
\end{IEEEbiography}
\vfill

\end{document}